\documentclass{article}

\usepackage{iclr2027_conference,times}
\usepackage{amsmath,amssymb}
\usepackage{graphicx}
\usepackage{booktabs}
\usepackage{multirow}
\usepackage{xcolor}
\usepackage[hidelinks]{hyperref}
\usepackage{url}



\newcommand{\AtmosAllGainCommon}{+11.7\%}
\newcommand{\AtmosAllGainRarest}{+92.2\%}
\newcommand{\AtmosAllP}{0.0583}
\newcommand{\AtmosAllRho}{-0.829}

\newcommand{\AvgCascOneEightyOneExact}{-0.0400}
\newcommand{\AvgCascOneEightyOnePre}{+0.0178}

\newcommand{\AvgCascOneSixtyExact}{-0.0304}
\newcommand{\AvgCascOneSixtyPre}{+0.0251}

\newcommand{\BootEvents}{936}
\newcommand{\BootHeadDiff}{+0.0339}
\newcommand{\BootHeadHi}{+0.0439}
\newcommand{\BootHeadLo}{+0.0234}

\newcommand{\BootHeadPRel}{<0.001}
\newcommand{\BootKnobDiff}{+0.1171}
\newcommand{\BootKnobHi}{+0.1246}
\newcommand{\BootKnobLo}{+0.1095}
\newcommand{\BootN}{2000}
\newcommand{\BootPMax}{< 0.001}

\newcommand{\CBootFlips}{51}

\newcommand{\CBootFlipsHi}{56}
\newcommand{\CBootFlipsLo}{45}
\newcommand{\CBootN}{2000}
\newcommand{\CBootRAll}{+0.796}
\newcommand{\CBootRAllHi}{+0.832}
\newcommand{\CBootRAllLo}{+0.753}

\newcommand{\CBootRMax}{+0.884}
\newcommand{\CBootRMaxHi}{+0.893}
\newcommand{\CBootRMaxLo}{+0.874}

\newcommand{\ContrastFlips}{51}
\newcommand{\ContrastFlipsBig}{23}
\newcommand{\ContrastFlipsBigThr}{0.01}
\newcommand{\ContrastFlipsNoGuid}{40}

\newcommand{\ContrastFlipsPub}{17}
\newcommand{\ContrastFlipsRelFive}{30}
\newcommand{\ContrastFlipsRelMed}{5.7\%}

\newcommand{\ContrastN}{450}
\newcommand{\ContrastNArms}{6}
\newcommand{\ContrastNPairs}{15}
\newcommand{\ContrastNPub}{90}
\newcommand{\ContrastNPubPairs}{3}
\newcommand{\ContrastRAll}{+0.796}
\newcommand{\ContrastRExt}{+0.869}
\newcommand{\ContrastRMax}{+0.884}
\newcommand{\ContrastRPub}{+0.791}

\newcommand{\ContrastRhoMax}{+0.787}

\newcommand{\CrowdBaseBiasMin}{-0.1286}

\newcommand{\CrowdBaseValMin}{-0.0151}

\newcommand{\CrowdCsrBiasMax}{+0.2162}

\newcommand{\CrowdCsrValMax}{+0.2394}

\newcommand{\CrowdFlatHardBias}{0.761}
\newcommand{\CrowdFlatHardGain}{+0.0809}

\newcommand{\CrowdN}{320}
\newcommand{\CrowdNAll}{1760}
\newcommand{\CrowdOrdN}{263}
\newcommand{\CrowdOrdOther}{57}
\newcommand{\CrowdOrdRho}{0.563}

\newcommand{\CrowdPostRho}{0.593}

\newcommand{\CrowdRho}{0.568}
\newcommand{\CrowdRhoAll}{0.629}
\newcommand{\CrowdSeedBiasHi}{1.146}
\newcommand{\CrowdSeedBiasLo}{0.002}
\newcommand{\CrowdSeedCsiHi}{0.3194}
\newcommand{\CrowdSeedCsiLo}{0.0017}
\newcommand{\CrowdSeedFold}{188}
\newcommand{\CrowdSeedMaeFold}{1.7}
\newcommand{\CrowdSeedMaeHi}{8.1}
\newcommand{\CrowdSeedMaeLo}{4.7}
\newcommand{\CrowdSeedMaeRho}{-1.000}

\newcommand{\CrowdSeedRho}{0.90}
\newcommand{\CrowdSeedTau}{10}
\newcommand{\CrowdSeeds}{5}
\newcommand{\CrowdShiftBiasHi}{0.979}
\newcommand{\CrowdShiftBiasLo}{0.926}

\newcommand{\CrowdShiftGain}{0.0023}

\newcommand{\DCFlips}{16}
\newcommand{\DCFlipsSep}{10}

\newcommand{\DCMaxSixteenCalRel}{+5.0\%}

\newcommand{\DCMaxSixteenRawRel}{+82.9\%}

\newcommand{\DCNLive}{28}

\newcommand{\DCRemovedHi}{97\%}
\newcommand{\DCRemovedLo}{73\%}
\newcommand{\DCReproGapHi}{0.046}
\newcommand{\DCReproGapLo}{0.011}

\newcommand{\DeflBiasM}{0.281}
\newcommand{\DeflBiasP}{1.020}
\newcommand{\DeflCal}{+31.3\%}

\newcommand{\DeflRaw}{+8.2\%}

\newcommand{\DoseCells}{30}
\newcommand{\DoseNAll}{150}
\newcommand{\DoseNArms}{5}
\newcommand{\DosePts}{30}
\newcommand{\DoseRho}{0.675}
\newcommand{\DoseRhoAll}{0.599}

\newcommand{\ECCsiSeventyTwoCal}{+0.0850}
\newcommand{\ECCsiSeventyTwoRaw}{+0.0502}

\newcommand{\ECN}{270}

\newcommand{\ECPssSeventyTwoCal}{+0.1167}
\newcommand{\ECPssSeventyTwoRaw}{+0.0245}
\newcommand{\ECR}{0.679}
\newcommand{\ECRho}{0.213}
\newcommand{\EFCascCal}{+18.9\%}
\newcommand{\EFCascErased}{60\%}
\newcommand{\EFCascRaw}{+47.4\%}
\newcommand{\EightAvgCasc}{0.2276}
\newcommand{\EightAvgDet}{0.2347}
\newcommand{\EightAvgNBetter}{0}
\newcommand{\EightCasc}{0.3069}
\newcommand{\EightCascBias}{1.037}
\newcommand{\EightCascGain}{-2.9\%}

\newcommand{\EightCascPub}{0.2841}
\newcommand{\EightCascPubDev}{+8.0\%}
\newcommand{\EightCascQ}{0.2978}
\newcommand{\EightDet}{0.1467}
\newcommand{\EightDetBias}{0.172}
\newcommand{\EightDetGain}{+79.8\%}
\newcommand{\EightDetK}{0.3201}
\newcommand{\EightDetQ}{0.2639}

\newcommand{\EightGapCal}{0.0339}
\newcommand{\EightGapRaw}{0.1601}
\newcommand{\EightGapReduction}{78.8\%}

\newcommand{\EightRelReduction}{88.2\%}
\newcommand{\EightReported}{+109.1\%}
\newcommand{\EightReportedAbs}{109.1\%}
\newcommand{\EightReportedPub}{+91.8\%}
\newcommand{\EightResidK}{-5.5\%}
\newcommand{\EightResidQ}{+12.9\%}
\newcommand{\EightResidQAbs}{12.9\%}
\newcommand{\EightShareHi}{88\%}
\newcommand{\EightShareK}{105\%}
\newcommand{\EightShareLo}{36\%}
\newcommand{\EightShareLoTau}{74}

\newcommand{\EightShareQ}{88\%}
\newcommand{\EoneAbsLo}{-0.0062}
\newcommand{\EoneAbsMax}{+0.0238}

\newcommand{\EoneCsiFold}{31}
\newcommand{\EoneExclHi}{0.0315}
\newcommand{\EoneExclLo}{0.0258}
\newcommand{\EoneExcluded}{4}
\newcommand{\EoneMax}{+250.8\%}

\newcommand{\EoneMin}{-2.3\%}
\newcommand{\EoneNRuns}{10}

\newcommand{\EoneP}{0.0028}
\newcommand{\EonePGraded}{0.0167}
\newcommand{\EonePTiered}{0.0583}
\newcommand{\EoneRho}{-1.000}
\newcommand{\EoneRhoGraded}{-0.943}
\newcommand{\EoneRhoTiered}{-0.829}
\newcommand{\EoneRmax}{20.270\%}
\newcommand{\EoneRmin}{0.264\%}
\newcommand{\EoneVariants}{graded and tiered}

\newcommand{\EtsErasedHi}{88\%}
\newcommand{\EtsErasedLo}{41\%}

\newcommand{\EtsMaxGap}{5}
\newcommand{\EtsNThr}{5}

\newcommand{\EtsThrNoGain}{16}
\newcommand{\ExactAvgFlips}{5}

\newcommand{\ExactAvgR}{0.874}

\newcommand{\ExactFlips}{64}

\newcommand{\ExactN}{450}

\newcommand{\ExactR}{0.896}
\newcommand{\ExactRefCal}{-0.0208}
\newcommand{\ExactRefCalCI}{[-0.0282, -0.0131]}

\newcommand{\ExactRefCalRel}{-6.5\%}
\newcommand{\ExactRefRaw}{+0.1615}

\newcommand{\ExactRefRemoved}{113\%}
\newcommand{\ExactRefShare}{106\%}

\newcommand{\ExactRho}{0.860}
\newcommand{\ExactSep}{46}
\newcommand{\GuidBias}{0.548}
\newcommand{\GuidCsi}{0.2475}
\newcommand{\GuidGain}{+17.1\%}

\newcommand{\KnobCsiLoseBias}{0.4030}
\newcommand{\KnobCsiLoseCsi}{0.3977}
\newcommand{\KnobCsiLosePool}{max $16^2$}
\newcommand{\KnobCsiLoseTau}{181}
\newcommand{\KnobDetKnob}{shift$=$32}

\newcommand{\KnobDetPerhCsi}{0.3413}
\newcommand{\KnobDetPooledBias}{0.3201}

\newcommand{\KnobDetUncal}{0.1467}

\newcommand{\KnobErasedPerhCsib}{109.9\%}

\newcommand{\KnobErasedPooledb}{105.4\%}
\newcommand{\KnobRaw}{+111.3\%}
\newcommand{\KnobRawb}{+107.6\%}

\newcommand{\KnobResidPerhCsi}{-9.6\%}
\newcommand{\KnobResidPerhCsib}{-10.7\%}
\newcommand{\KnobResidPooled}{-4.4\%}

\newcommand{\KnobResidPooledb}{-5.8\%}

\newcommand{\LeadErasedG}{87.3\%}
\newcommand{\LeadErasedGb}{89.1\%}
\newcommand{\LeadErasedLe}{76.4\%}
\newcommand{\LeadErasedLeb}{76.7\%}
\newcommand{\LeadErasedLm}{84.8\%}
\newcommand{\LeadErasedLmb}{86.3\%}
\newcommand{\LeadGainG}{+79.8\%}
\newcommand{\LeadGainLe}{+62.1\%}
\newcommand{\LeadGainLm}{+74.1\%}

\newcommand{\LeadRaw}{+111.3\%}
\newcommand{\LeadRawb}{+107.6\%}
\newcommand{\LegCfgLin}{57\%}
\newcommand{\LegCfgLog}{33\%}
\newcommand{\LegDiffLin}{43\%}
\newcommand{\LegDiffLog}{67\%}

\newcommand{\PDMoveMax}{0.020}

\newcommand{\PDNLive}{29}

\newcommand{\PDNearOne}{28}

\newcommand{\PDShareMax}{4\%}

\newcommand{\PaperFamAvgFlips}{14}
\newcommand{\PaperFamAvgN}{180}
\newcommand{\PaperFamAvgR}{0.302}

\newcommand{\PaperFamMaxN}{270}
\newcommand{\PaperFamMaxR}{0.889}

\newcommand{\PersBiasHi}{1.020}
\newcommand{\PersBiasLo}{0.993}
\newcommand{\PersGainMax}{0.8\%}
\newcommand{\QmapEvents}{33}
\newcommand{\QmapPixels}{5.8\times10^{7}}
\newcommand{\QmapQuantiles}{1001}

\newcommand{\RareTop}{0.081\%}
\newcommand{\RareTopOne}{1238}

\newcommand{\ReproWorst}{2.8\%}
\newcommand{\RerunFlips}{50}
\newcommand{\RerunR}{0.801}

\newcommand{\SameArchBiasA}{0.281}
\newcommand{\SameArchBiasB}{0.172}
\newcommand{\SameArchCal}{+5.3\%}
\newcommand{\SameArchExactFlips}{6}
\newcommand{\SameArchExactN}{6}
\newcommand{\SameArchFlips}{5}
\newcommand{\SameArchN}{6}
\newcommand{\SameArchRaw}{-29.5\%}
\newcommand{\SegArch}{deeplabv3-resnet50}

\newcommand{\SegBestCls}{boat}
\newcommand{\SegBestGain}{+0.0272}
\newcommand{\SegCtrBias}{0.963}
\newcommand{\SegCtrCls}{tvmonitor}
\newcommand{\SegCtrGain}{-0.0248}
\newcommand{\SegCtrGainVal}{-0.0101}
\newcommand{\SegCtrR}{0.414\%}

\newcommand{\SegDiffA}{all val2017 images, not the VOC-containing subset}
\newcommand{\SegDiffB}{iscrowd regions kept in the target}

\newcommand{\SegFlatThr}{0.001}

\newcommand{\SegMeanGain}{-0.0020}

\newcommand{\SegNCls}{20}

\newcommand{\SegNearFlat}{6}

\newcommand{\SegNearHelped}{1}
\newcommand{\SegNearHurt}{4}
\newcommand{\SegNearMeanGain}{-0.0048}
\newcommand{\SegNearOne}{11}
\newcommand{\SegNearThr}{0.06}

\newcommand{\SegOverCls}{diningtable}
\newcommand{\SegOverGain}{-0.0351}

\newcommand{\SegReproDev}{-5.3\%}

\newcommand{\SegRhoBias}{+0.663}
\newcommand{\SegRhoBiasPoolA}{+0.836}

\newcommand{\SegRhoRarity}{-0.019}
\newcommand{\SixBiasMaxFour}{0.214}
\newcommand{\SixBiasMaxSixteen}{0.172}
\newcommand{\SixBiasNone}{0.291}
\newcommand{\SixGainHi}{+79.8\%}
\newcommand{\SixGainLo}{-8.9\%}
\newcommand{\SixGainMaxFour}{+61.1\%}
\newcommand{\SixGainMaxSixteen}{+79.8\%}
\newcommand{\SixGainNone}{+39.3\%}
\newcommand{\SixPoolHi}{max $16{\times}16$}
\newcommand{\SixPoolLo}{average $16{\times}16$}

\newcommand{\SixRevMaxSixteen}{5}
\newcommand{\SixRevOther}{0}

\newcommand{\SixSpread}{89}
\newcommand{\TTDCInside}{81\%}

\newcommand{\TTDCMaxG}{4.5}
\newcommand{\TTDCN}{84}
\newcommand{\TTDCNBig}{65}
\newcommand{\TTDCNInRange}{59}

\newcommand{\TTDCRho}{0.92}
\newcommand{\TTDCSign}{97\%}
\newcommand{\TTIntercept}{0.0015}
\newcommand{\TTPDInside}{100\%}
\newcommand{\TTPDMAE}{0.003}

\newcommand{\TTPDN}{29}

\newcommand{\TTResidSD}{0.0176}
\newcommand{\TTSlope}{0.086}
\newcommand{\ValCascBias}{1.037}
\newcommand{\ValCascCloser}{2.7}
\newcommand{\ValCascTest}{-2.9\%}
\newcommand{\ValCascVal}{-0.2\%}
\newcommand{\ValCells}{450}
\newcommand{\ValErasedTest}{88\%}
\newcommand{\ValErasedVal}{96\%}
\newcommand{\ValFlatArms}{persistence and pysteps}
\newcommand{\ValFlatMax}{0.8\%}
\newcommand{\ValFlatOtherBias}{0.907}
\newcommand{\ValFlipsTest}{51}
\newcommand{\ValFlipsVal}{49}
\newcommand{\ValHeadRaw}{+109.1\%}
\newcommand{\ValHeadTest}{+12.9\%}
\newcommand{\ValHeadVal}{+4.6\%}
\newcommand{\ValN}{128}

\newcommand{\ValNPay}{3}
\newcommand{\ValRTest}{+0.796}
\newcommand{\ValRVal}{+0.766}
\newcommand{\ValRatioHi}{188\%}
\newcommand{\ValRatioHiArm}{EarthFormer}
\newcommand{\ValRatioLo}{44\%}
\newcommand{\ValRatioLoArm}{CasCast cascade (no guidance)}
\newcommand{\WBFlips}{49}
\newcommand{\WBFlipsSep}{41}
\newcommand{\WBFuxiBias}{0.16}
\newcommand{\WBFuxiCal}{+0.0819}

\newcommand{\WBFuxiRaw}{-0.0248}

\newcommand{\WBN}{360}
\newcommand{\WBR}{0.817}

\newcommand{\WBRho}{0.871}

\iclrfinalcopy
\hypersetup{
  pdftitle={A Free Knob: A Sensitivity Audit for Calibration in Threshold-Based Evaluation},
  pdfauthor={Md Tanveer Hossain Munim, Bijoy Ahmed Saiem, Al-Amin Sany, Tanzima Hashem}}
\newcommand{\CodeURL}{\url{https://github.com/munim110/free-knob}}

\newcommand{\csi}{\mathrm{CSI}}
\newcommand{\bias}{\mathrm{Bias}}

\title{A Free Knob: A Sensitivity Audit for Calibration in Threshold-Based Evaluation}

\author{Md Tanveer Hossain Munim$^{1,2}$, Bijoy Ahmed Saiem$^{2}$, \\
\textbf{Al-Amin Sany$^{2}$ \& Tanzima Hashem$^{2}$} \\
$^{1}$Regional Integrated Multi-Hazard Early Warning System (RIMES) \\
$^{2}$Bangladesh University of Engineering and Technology (BUET) \\
\texttt{\{1705057,1905052,1905048\}@ugrad.cse.buet.ac.bd} \\
\texttt{tanveer.munim@outlook.com}, \texttt{tanzimahashem@cse.buet.ac.bd}}

\begin{document}
\maketitle
\lhead{Preprint}
\raggedbottom

\begin{abstract}
Many benchmarks for rare events score a prediction in three steps: pool the
prediction and the target over spatial blocks, threshold each pooled field, and
score the resulting contingency table. We show that at a fixed rare threshold
the max-pooled Critical Success Index (CSI) cannot separate two different
things: how well a system places events, and how well its amplitudes are
calibrated. Sharp observations push many blocks above the threshold, while the
attenuated predictions of squared-error regression leave those same blocks
below it. We turn a classical remedy into a test. A monotone calibration map is
fitted on held-out data and applied separately to each system. The map needs no
retraining and cannot change the order of any two pixels, so whatever part of a
difference it reproduces cannot be evidence of better spatial ranking. We report
how much of a reported difference the map removes. On SEVIR, two released
checkpoints of one architecture differ by \SameArchRaw\ in extreme-threshold CSI
before the control and by \SameArchCal\ after it. Across \ContrastN\ pairwise
contrasts among \ContrastNArms\ systems, how far apart two systems sit in pooled
frequency bias predicts how far their CSI difference moves under the control
($r=\ExactR$ when the map is fitted after pooling), and \ExactFlips\ contrasts
reverse sign. At CasCast's published extreme-event operating point, calibration
alone removes \EightShareQ--\ExactRefShare\ of the cascade's gain over its
backbone across three controls. The effect
persists when the map is fitted on a window preceding the test period, and it
acts in both directions. Calibration also reveals advantages hidden by a
better-calibrated baseline. We also test when the problem is largest. On
geostationary infrared imagery, a larger share of the score can be explained by
calibration as events get rarer. Crowd-counting experiments, which use sum
pooling instead of max pooling, show the same link between a system's frequency
bias and how much calibration shifts its score. Semantic segmentation is the
opposite case. There, frequency bias is already close to one, hence calibration
changes almost nothing. The problem therefore needs two things at the same time: a threshold that is
fixed in advance, and a model whose output is poorly calibrated at that
threshold. We recommend reporting pooled frequency bias and a symmetric held-out
\emph{FreeKnob Audit}, released as an installable package, alongside rare-event
pool-and-threshold scores.
\end{abstract}

\section{Introduction}

Many benchmarks for rare, high-magnitude events score a prediction in three
steps \citep{gong2024cascast,song2026extreme,nguyen2025raindiff}. First, the
predicted and the observed field are pooled over spatial blocks, so that small
errors in location are tolerated. Second, each pooled field is thresholded,
which picks out the events of interest. Third, the hits, misses, and false
alarms are counted. In precipitation nowcasting the usual score is the Critical
Success Index (CSI): hits divided by hits plus misses plus false alarms.

CSI does not measure placement alone. It also depends on \emph{frequency bias},
the ratio of predicted to observed event counts \citep{schaefer1990csi}. A
system that predicts too few events loses score through misses, even when every
event it does predict is in the right place. A system that predicts more events
can raise its score without placing any of them better. The comparisons we audit
report pooled CSI without first equalising how many events each system predicts.
This paper asks one question: \emph{how much of a reported CSI difference
between two systems can be produced by calibrating amplitudes alone, without
placing events any better?}

\begin{figure}[t]
\centering
\includegraphics[width=0.88\textwidth]{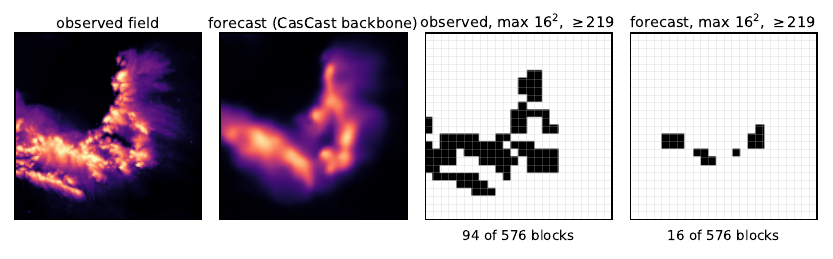}
\caption{One SEVIR event. The forecast follows the observed storm, but its peaks
are too low. After $16{\times}16$ max pooling and thresholding at $219$, the
observation has 94 positive blocks and the forecast has 16, so the forecast is
missing most of the blocks it should have. The ratio $16/94\approx0.17$ is
closer to the backbone's pooled frequency bias over the whole split
(\EightDetBias) than that of any other frame, so this event shows the deficit
the backbone usually has.}
\label{fig:illustration}
\end{figure}

The question is particularly important on SEVIR \citep{veillette2020sevir}, a
storm-event radar dataset widely used to benchmark deep precipitation nowcasting. At the
highest commonly reported SEVIR threshold, only \RareTop\ of pixels are
positive, about one in every \RareTopOne. CasCast \citep{gong2024cascast} adds a
diffusion cascade on top of a deterministic backbone, and reports its main
extreme-event comparison at that threshold using $16{\times}16$ max pooling.
More recent generative nowcasting studies report the same score
\citep{song2026extreme,nguyen2025raindiff}.

Max pooling treats a sharp observation and a smooth prediction differently. A
block counts as positive if any single pixel inside it crosses the threshold, so
a sharp observed storm turns many blocks positive while a smoother prediction of
the same storm leaves some of them just below it (Figure~\ref{fig:illustration}).
Those blocks are counted as misses, which pushes the pooled frequency bias below
one, even when the prediction ranks its pixels exactly as the observation does
(Section~\ref{sec:mechanism}).

Fixing this frequency-bias deficit changes two comparisons of released models. The
first involves two checkpoints of EarthFormer's cuboid-attention architecture
\citep{gao2022earthformer}, trained independently. At this threshold they differ
by \mbox{\SameArchRaw} in CSI. A gap of this size would ordinarily be read as an
architectural improvement, yet the two checkpoints share the same architecture.
After the same held-out calibration is applied to each of them, the difference
becomes \SameArchCal, and the sign reverses at \SameArchFlips\ of the
\SameArchN\ thresholds (Section~\ref{sec:samearch}). The second is the CSI gain
of the CasCast cascade over its own deterministic backbone. Across three
controls, calibration alone removes \EightShareQ--\ExactRefShare\ of it
(Section~\ref{sec:cascast}). Calibrating amplitudes alone therefore changes a
CSI comparison between released models by a large amount. Rankings like these
guide which architectures the field builds on, so a calibration difference read
as a modelling advance misdirects that choice.

We measure how much of a comparison calibration alone can account for, using a
monotone transform fitted on held-out data. The transform shifts amplitudes,
rescales them, or maps the forecast's distribution of values onto the observed
one. It needs no retraining, and it cannot swap the order of any two pixels. We
call it a \emph{free knob}. It changes how much forecast mass crosses the
threshold, but it cannot improve where events are placed. Under max pooling, the
quantile-map version is matched-percentile thresholding, the classical remedy
\citep{mittermaier2013long}, up to how empirical ties are handled. We use that
remedy as a symmetric control. The same calibration procedure is fitted
separately to every system evaluated, which we call an \emph{arm}. We then
ask how much of the original difference is left, and call that the \emph{residual}.

This reveals a concrete identification failure. \emph{At a fixed rare threshold,
pool-and-threshold CSI cannot tell spatial discrimination apart from amplitude
calibration.} A difference that disappears under this control does not, by
itself, establish improved spatial ranking. A difference that survives is what
remains after the stated calibration. It is not a direct measurement of spatial
skill or a guaranteed bound on it. We therefore report the share of the difference
the control removes. In Section~\ref{sec:cascast} that share is at least \EightShareQ\ under each of three controls, while the sign of the residual depends on how much calibration each
arm is allowed. Pooled frequency bias is the
diagnostic. If two arms are evaluated at very different frequency biases, the
raw CSI difference mixes the intended comparison with that operating-point
difference.

Across \ContrastN\ pairwise contrasts among \ContrastNArms\ arms, how far apart two
arms sit in frequency bias tracks how far their CSI difference moves under the
control, and \ExactFlips\ of those contrasts reverse sign
(Section~\ref{sec:contrasts}). Calibration can shrink a claimed gain, and it can
also uncover one that a better-calibrated baseline was hiding. Further
experiments look at pooling, rarity, and other domains. Crowd counting reproduces
the bias--gain relationship. Segmentation shows small average movement where
frequency bias is near one (Section~\ref{sec:general}).

\paragraph{Contributions.} We (i) show how max pooling turns low forecast
amplitudes into missed events, and how that effect grows as events get rarer and
blocks get larger; (ii) decompose published comparisons on released
checkpoints, and show that differences in pooled frequency bias track how far
each comparison moves, across all \ContrastN\ contrasts, \ExactFlips\ of which
reverse sign; and (iii) show that the same effect appears in crowd counting under
sum pooling, and is small on average in the segmentation control, where
frequency bias is already near one. The resulting \emph{FreeKnob Audit}, shipped
as an installable package at \CodeURL, reports pooled frequency bias and
repeats every comparison after the same held-out calibration is fitted
separately to each arm. To our knowledge, it is the first such audit of
published deep-learning nowcasting comparisons.

\section{Related work}
\label{sec:related}

\paragraph{Frequency bias and verification.} CSI (the threat score) has long
been known to depend on frequency bias as well as discrimination
\citep{schaefer1990csi,hamill2006climatology}. Verification work also
studies the resulting incentives to hedge \citep{stephenson2000odds,hogan2010equitability}
and has used probability/frequency matching to compare forecasts at matched event
frequencies \citep{ebert2001pm,mittermaier2013long}, including bias
equalisation before numerical models are compared
\citep{hamill1999hypothesis,pyle2019comparison}. Frequency matching has also
been applied to a deep-learning nowcast as bias correction
\citep{zhang2022frequency}. We establish how strongly this
known sensitivity changes comparisons of released deep nowcasting systems at
rare, max-pooled operating points.

\paragraph{Pooling and alternative scores.} Spatial pooling belongs to the
neighbourhood-verification family
\citep{ebert2008fuzzy,gilleland2009intercomparison}. Earlier work showed that
fixed-threshold neighbourhood verification can confound intensity error with
spatial or size error, and proposed percentile-based thresholds to reduce that
effect \citep{zhu2015percentile}. We differ in using a held-out, symmetric marginal
control on released checkpoints while retaining the original CSI operating
point. Related work also shows that neighbourhood scores such as FSS are
themselves sensitive to frequency bias
\citep{necker2024pfss}, and that rankings of AI and numerical precipitation
forecasts depend on neighbourhood size \citep{loveday2025extreme}. Under max pooling, the quantile-map control corresponds to
matched-percentile thresholding, subject to empirical tie and interpolation
conventions (Section~\ref{sec:setup}). Replacing CSI with another dichotomous
score does not remove the problem. Bias-adjusted, chance-corrected, and
rare-event alternatives make different assumptions and can remain inequitable at
low base rates
\citep{mesinger2004bias,hogan2010equitability,stephenson2008eds,ferro2011edi}.
Reweighting the contingency table cannot supply our control either, because
pooling and thresholding have already fixed which samples enter it. We therefore
transform the forecast and re-score the same metric. Appendix~\ref{app:relwork}
reviews these alternatives and Appendix~\ref{app:ets} repeats the main
decomposition under ETS.

\paragraph{Recalibration and benchmark audits.} Monotone recalibration is
standard in machine learning \citep{platt1999probabilistic,guo2017calibration},
and the broader distinction between calibration and resolution is classical
\citep{murphy1987decomposition,gneiting2007sharpness}. Recent
forecast-verification work likewise argues for fair comparison by granting
competing deterministic weather models the same statistical post-processing and
evaluating their potential CRPS \citep{gneiting2025potential}, a construction
since extended to extremes with weighted potential CRPS
\citep{biegert2026weighted}. There, model ordering changes little when different
extreme-event thresholds are emphasised. That work asks a different question
than we do. It checks whether rankings stay stable when the weight given to
extreme thresholds changes in a post-processed score. We check whether
the gap between two models at a single fixed threshold changes once each model
is recalibrated. Our transform is a negative control. It does not replace
the benchmark score. Relatedly, \citet{ranjan2024posthoc} show that model-performance trends can
reverse after post-hoc transformations and argue that model-selection conclusions
should account for those transformations. This places the paper alongside
benchmark audits showing that apparent progress can depend on unequal treatment
of compared systems
\citep{musgrave2020reality,dacrema2019troubling,henderson2018deeprl,bello2021revisiting}.
Here the control requires no retraining or hyperparameter search, and its effect
is diagnosed by pooled frequency bias.

\paragraph{Sharpness and pooled CSI.} Modern nowcasting papers often read
high pooled CSI as evidence that predictions are sharper or capture local
extremes better. This interaction has been observed directly. FACL reports that
for sharp forecasts CSI increases with pooling size, whereas for blurry
forecasts it changes little with pooling size \citep{yan2024facl}. Their
experiment changes the training loss. Ours holds released predictions fixed and
grants each the same held-out marginal control. Sharpness can be valuable, but fixed-threshold pooled CSI need not
separate getting amplitudes across the threshold from putting those amplitudes
in the right places. Reporting practice varies. RainPro-8 tabulates frequency
bias by threshold and finds that EarthFormer and SimVP substantially
under-predict \citep{sarabia2026rainpro}, whereas the SEVIR comparisons audited
here report pooled CSI without it.

\section{Method: the FreeKnob Audit}
\label{sec:setup}

\paragraph{Metric.} For a forecast field $\hat y$, observation $y$, threshold
$\tau$, and pooling operator $\Pi$, let $h,f,m$ be hits, false alarms, and misses
after pooling both fields and thresholding:
\begin{equation}
\begin{aligned}
h &= \textstyle\sum \mathbf{1}[\Pi\hat y\ge\tau]\mathbf{1}[\Pi y\ge\tau],\\
f &= \textstyle\sum \mathbf{1}[\Pi\hat y\ge\tau]\mathbf{1}[\Pi y<\tau],\\
m &= \textstyle\sum \mathbf{1}[\Pi\hat y<\tau]\mathbf{1}[\Pi y\ge\tau].
\end{aligned}
\end{equation}
We report $\csi=h/(h+f+m)$ and pooled frequency bias
$\bias=(h+f)/(h+m)$, accumulating one contingency table over the split. A
\emph{cell} is one $(\Pi,\tau)$ pair. The \emph{reference cell} is max
$16{\times}16$ pooling at $\tau=219$, where CasCast reports its extreme-event result.

\paragraph{The free knob.} Let $g$ be an order-preserving amplitude map applied
to the raw forecast before pooling and thresholding. We consider shifts, positive
gains, and marginal quantile mapping,
\begin{equation}
\mathcal G=\{x\mapsto x+\delta\}\cup\{x\mapsto\gamma x:\gamma>0\}
\cup\{x\mapsto F_y^{-1}(F_{\hat y}(x))\}.
\end{equation}
Such a map cannot reverse the ordering of any two pixels, so it cannot improve
spatial ranking. It only changes which amplitudes cross the operating threshold. We call the
resulting change in CSI the \emph{knob gain}.
The observed marginal
used to fit it carries information about amplitude but no paired information
about location. Throughout, ``calibration'' means this marginal,
operating-point notion, summarised by pooled frequency bias rather than
probabilistic reliability. We use $\mathrm{dev}=|\log\bias|$ as a symmetric
measure of deviation from $\bias=1$. We use $r$ for Pearson correlation and
$\rho$ for Spearman rank correlation.

For continuous, strictly increasing marginals,
$g(\hat y)\ge\tau\iff\hat y\ge F_{\hat y}^{-1}(F_y(\tau))$, which is
matched-percentile thresholding \citep{mittermaier2013long}. Our empirical map
uses linear interpolation and explicit tie conventions (Appendix~\ref{app:reproducibility}).
A monotone map commutes with max pooling but need not with average pooling.
Under average or sum pooling a nonlinear monotone map applied before pooling can
change the order of two pooled blocks. We therefore also report an \emph{exact
control}, which fits the same map \emph{after} pooling, on the pooled field that
CSI scores rather than on pixel amplitudes. As a result, it cannot change the ranking of
pooled cells under any convention. Under no pooling and under max pooling the
two are both per-arm thresholds and differ only in how the threshold is
matched. Under average pooling they calibrate different objects, so we report
both. The \emph{FreeKnob Audit} applies the same procedure separately to every
arm on held-out data, reports the share of the original contrast the control
removes, and places pooled frequency bias beside the raw score.

\paragraph{Fitting protocol.} We split events by index parity into calibration
and evaluation halves. All main results use one \emph{global} quantile map
per arm, fitted once on \QmapQuantiles\ quantiles and then held fixed across
every threshold and pooling convention. The fit uses \QmapEvents\ evenly spaced
calibration events ($\QmapPixels$ pixels), the largest sample under our fixed
pixel cap. Refitting on \ValN\ events from an earlier window leaves the
conclusions unchanged (Section~\ref{sec:cascast}). We also
report a stronger \emph{per-cell} variant, selected separately for each
$(\Pi,\tau)$ to minimise $|\bias-1|$ on the calibration half. Its role is to show
how conclusions depend on the allowed calibration budget
(Section~\ref{sec:cascast} and Appendix~\ref{app:pooling}). Neither variant is selected on evaluation CSI. The
global map is fitted to the raw marginal, so it does not force post-pooling
$\bias=1$. Section~\ref{sec:cascast} additionally fits it on a window preceding
the test period.

\section{Experimental setup}
\label{sec:expsetup}

\paragraph{Data and models.} We use SEVIR \citep{veillette2020sevir}: 1872 test
events, 13 input frames, 12 output frames at $384{\times}384$, and VIL thresholds
$\{16,74,133,160,181,219\}$. The \DoseNArms\ primary arms are persistence
(the last observed frame repeated at every lead time),
pysteps Lucas--Kanade advection \citep{pulkkinen2019pysteps}, the official
EarthFormer checkpoint \citep{gao2022earthformer}, and CasCast's released
deterministic backbone and diffusion cascade \citep{gong2024cascast}. All learned arms are released artefacts.

\paragraph{Reproduction fidelity.} The central experiment decomposes a published
number, so matching the released evaluation protocol matters. We run the
released checkpoints through one common pipeline and compare our deterministic
CasCast backbone directly with the authors' reported cells. The worst discrepancy
is \ReproWorst. One implementation detail affects the scores. CasCast's
released evaluation script samples with classifier-free guidance weight $2$.
Using the unguided path, although it is a natural reading of the model
description, gives a different extreme-threshold score. We
therefore use the released guided setting throughout and keep the unguided sample
only as a sixth diagnostic arm. Appendix~\ref{app:repro} records the
checks and Appendix~\ref{app:guid} isolates the guidance effect.

\section{The mechanism and its effect size}
\label{sec:mechanism}

Consider a forecast that places precipitation perfectly but reproduces only a
fraction of the observed local dynamic range, a characteristic failure mode of
squared-error regression under uncertainty. The population minimiser of squared error is the
conditional mean, which averages over sharp outcomes and can therefore be
smooth. Without pooling, a smoothed forecast predicts peaks that are too low, so any
pixel whose observed value is above the threshold but whose forecast falls
below it is counted as a miss. Under max pooling with block size $k$ it is
compared against $\max_{k^2} y$, an order statistic that rises with $k$ for the
sharp observation and much less for the smooth forecast, so the gap widens with
$k$ and with $\tau$ and enters the contingency table as misses, driving $\bias$
below one.

Persistence is a control because its forecasts are observed, sharp radar
fields. If the pooled marginal is approximately stable over
the forecast window, its bias should stay near one and the knob should move its
score little. Storm growth or decay can violate that condition.

\begin{figure}[t]
\centering
\includegraphics[width=0.80\textwidth]{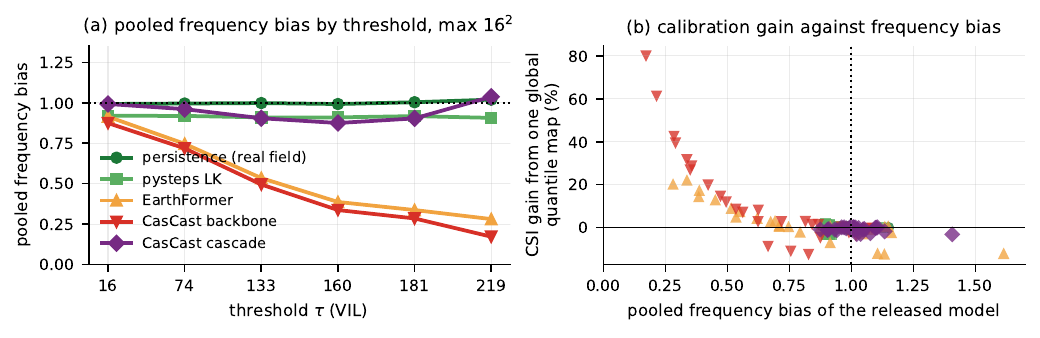}
\caption{\textbf{(a)} Pooled frequency bias across thresholds under max
$16{\times}16$ pooling. \textbf{(b)} The consequence: each of the \DosePts\ points is one arm at one threshold under max
$16{\times}16$, the vertical axis is the CSI gain from a
single global quantile map fitted on held-out events. The knob gain grows with
the bias deficit and is generally small near $\bias = 1$ in the evaluated arms. The event in
Figure~\ref{fig:illustration} is the frame whose block ratio is closest to the
backbone's pooled bias over the whole split.}
\label{fig:mech}
\end{figure}

As predicted, persistence's pooled bias stays within
$[\PersBiasLo, \PersBiasHi]$ across all six thresholds under max $16{\times}16$
pooling, and the global quantile map changes its CSI by at most \PersGainMax, in
some cells slightly negative, as expected for a transform fitted on a different
half. Figure~\ref{fig:mech}b plots knob gain against bias for the \DosePts\
arm--threshold points the \DoseNArms\ primary arms give at max $16{\times}16$,
the reference convention. The gain is ordered by how far each arm's bias
sits below one, $\rho = \DoseRho$, and
$\rho = \DoseRhoAll$ over all \DoseNAll\ cells once the other four pooling
conventions are included as well. Both correlations are limited by ties.
Three of the \DoseNArms\ arms sit at $\bias \approx 1$ with gains within a
percentage point of zero, and their relative order is
noise. The same holds for the strongest arm. CasCast's cascade has the highest CSI we score, and its bias is already
\EightCascBias\ at $\tau = 219$. The map that lifts its own backbone by
\EightDetGain\ moves it by \EightCascGain, so calibration adds
little once sharpness is restored.

\paragraph{Rarity.}
\label{sec:rarity}
The released SEVIR checkpoints are scored at thresholds they were not trained
for, and retraining them at each threshold is not feasible. Rarity is therefore
varied on a smaller task where the model is retrained at every operating point. The task is downscaling ERA5 reanalysis predictors \citep{hersbach2020era5} to infrared
brightness temperature from the geostationary Himawari-8 satellite
\citep{bessho2016himawari}, with the event defined as a temperature at or
below $\tau$ (Appendix~\ref{app:atmos}). We retrain an attention U-Net at each of six
operating points whose positive fraction ranges from $R=\EoneRmax$ to
$R=\EoneRmin$. The knob gain rises from \EoneMin\ at the most
common level to \EoneMax\ at the rarest, with monotone ordering
($\rho=\EoneRho$ against $\log_{10}R$). The absolute CSI change remains bounded
within $[\EoneAbsLo,\EoneAbsMax]$ while CSI itself collapses
\EoneCsiFold-fold, so calibration becomes an increasingly large fraction of the
reported score as events become rare (full sweep and permutation tests in
Appendix~\ref{app:rarity}).

\paragraph{Pooling.}
\label{sec:pooling}
The pooling convention also changes the knob gain. For CasCast's deterministic
backbone at $\tau=219$, it is \SixGainLo\ under
\SixPoolLo\ but \SixGainHi\ under \SixPoolHi, an \SixSpread-point range with the
model held fixed. The deficit deepens with max-pooling block size. The
backbone's bias falls from \SixBiasNone\ unpooled to \SixBiasMaxFour\ at
$4{\times}4$ and \SixBiasMaxSixteen\ at $16{\times}16$, and the gain rises from
\SixGainNone\ to \SixGainMaxFour\ and \SixGainMaxSixteen. Under max $16{\times}16$ pooling, pysteps outranks EarthFormer
at \SixRevMaxSixteen\ of 6 thresholds. Under every other convention it does so
at \SixRevOther\ (full results in Appendix~\ref{app:pooling}).

\section{Decomposing a published comparison}
\label{sec:cascast}

\begin{figure}[t]
\centering
\includegraphics[width=0.88\textwidth]{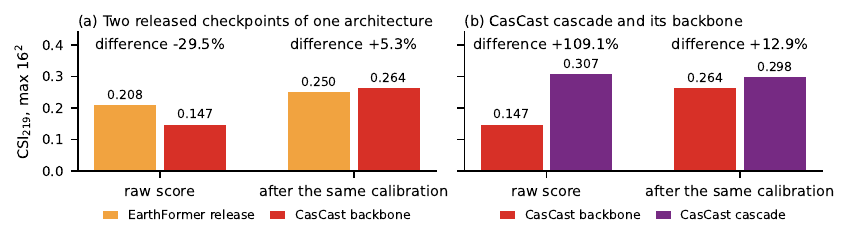}
\caption{Two comparisons of released SEVIR models at $\tau=219$ under max
$16{\times}16$ pooling, before and after the same held-out monotone calibration
is applied to each model. \textbf{(a)} Two independently trained checkpoints of
one architecture. \textbf{(b)} A published cascade against its backbone. The
calibration cannot reorder pixels, so the part of a raw difference that it
removes cannot be attributed to improved spatial ranking.}
\label{fig:consequence}
\end{figure}

CasCast \citep{gong2024cascast} reports that adding its diffusion cascade on top
of its own deterministic backbone improves $\csi_{219}$ under $16{\times}16$
max pooling by \EightReportedPub, and quotes this as evidence for cascaded
modelling of regional extremes. Both stages are released checkpoints, so the
comparison can be decomposed directly. We run both through one
pipeline on one split and report the $2{\times}2$ of $\{\text{backbone},
\text{cascade}\} \times \{\text{uncalibrated}, +\text{global quantile map}\}$. Our preprocessed SEVIR subset differs in sample and window construction from the
original CasCast evaluation, so agreement with the published scores is a
fidelity check. It does not reproduce that protocol exactly. The calibration is the same procedure in both arms, one map per arm fitted
on events neither is scored on. A single shared map would not be symmetric,
because it favours the arm whose marginal it was fitted on.

At the reference cell, the backbone scores \EightDet\ at pooled bias
\EightDetBias\ and the cascade \EightCasc\ at \EightCascBias. After symmetric
global calibration, the corresponding CSI scores are \EightDetQ\ and \EightCascQ\ (Figure~\ref{fig:consequence}b).
Thus, in our common-pipeline evaluation, the absolute cascade-over-backbone
gap falls from \EightGapRaw\ to \EightGapCal, a \EightGapReduction\ reduction.
Expressed instead as a relative advantage over the backbone, the contrast falls
from \EightReportedAbs\ to \EightResidQAbs, an \EightRelReduction\ reduction.
Each reduction is one minus the calibrated-to-raw ratio and
is reported only where the raw advantage is positive.

Our raw relative gain exceeds the published \EightReportedPub\ because our
cascade reproduction scores above the published value
(\EightCasc\ against \EightCascPub, \EightCascPubDev). We therefore treat the
published percentage as context rather than as the denominator of the
calibrated decomposition.

Beyond the reference cell, the
relative-gain reduction increases with threshold, from \EightShareLo\ at
$\tau=\EightShareLoTau$ to \EightShareHi\ at $\tau=219$, a threshold dependence consistent with the rarity
pattern of Section~\ref{sec:rarity} without identifying rarity separately from
intensity. Under the global map the residual is positive at
every threshold. Under the exact control it is negative at $\tau=74$, $133$
and $219$ (Table~\ref{tab:exactcascast}). Its sign therefore depends on the
control. The result is also specific to the reporting convention. Under
$16{\times}16$ average pooling the cascade leads at \EightAvgNBetter{} of 6
thresholds, and at $\tau=219$ scores \EightAvgCasc\ against the backbone's
\EightAvgDet.

\paragraph{Fitting before the evaluation period.} The global map is fitted on
the disjoint calibration half of the test split because that is the symmetric
control for a between-arm comparison. To test a
more deployment-like separation, we refit each arm on \ValN\ events from
1 January to 30 April 2019, which precedes the test period. The relative-gain reduction
is then \ValErasedVal\ rather than \ValErasedTest, and the calibration-gap
correlation is $r=\ValRVal$ rather than $r=\ValRTest$
(Appendix~\ref{app:valfit}).

The residual also depends on the admissible calibration budget. The
global map is restrictive by design. One transform per arm is fixed across all
thresholds and pooling conventions. A per-cell control, still fitted only on
held-out data and granted to both arms, removes \EightShareK\ of the raw gain
and leaves a residual of \EightResidK, which places the cascade behind its
backbone. The exact control of Section~\ref{sec:setup} removes
\ExactRefShare\ of the raw gain and leaves a residual of \ExactRefCalRel. It is scored on the re-run of Section~\ref{sec:contrasts}, a new sample from the diffusion cascade, where the CSI difference goes from \ExactRefRaw\ (against \EightGapRaw\ above) to \ExactRefCal\ \ExactRefCalCI. The share removed is therefore \EightShareQ--\ExactRefShare\ across the three controls. The sign of the residual changes with the budget. The
global map remains the primary control. Guidance and alternative knob-family
analyses are in Appendices~\ref{app:guid} and~\ref{app:knobsevir}.

\subsection{All pairwise contrasts}
\label{sec:contrasts}

To extend the decomposition beyond one paper, we run the same $2{\times}2$ over
every pair of the \ContrastNArms\ arms, \ContrastNPairs\ unordered pairs at
\DoseCells\ cells each, \ContrastN\ in all. These are repeated comparisons among
the same arms, not independent model replications. For a pair $(A,B)$ the
\emph{raw} contrast $\csi_B - \csi_A$ is what a results table prints, the
\emph{calibrated} contrast is the same difference after one global quantile map
fitted per arm is applied to both, and their difference is the distortion. That
distortion equals the difference of the two arms' knob gains, so it is not
independent evidence for the mechanism. It gives a prediction we can check. The
more two arms differ in how far each sits from unbiased, measured by the pair's
\emph{calibration gap} $\mathrm{dev}(A) - \mathrm{dev}(B)$, the more the control
should move the difference between them. The test could fail even if the
per-arm mechanism held, since two arms miscalibrated in the same direction
cancel. The prediction holds. The correlation is $r = \ContrastRAll$ over all \ContrastN\ contrasts,
$r = \ContrastRMax$ at the reference pooling, and $r = \ContrastRExt$ at the
extreme thresholds, with Spearman $\rho = \ContrastRhoMax$ at the reference
pooling against $\rho = \DoseRho$ for the single-arm relationship of
Section~\ref{sec:mechanism}.

\begin{figure}[t]
\centering
\includegraphics[width=0.86\textwidth]{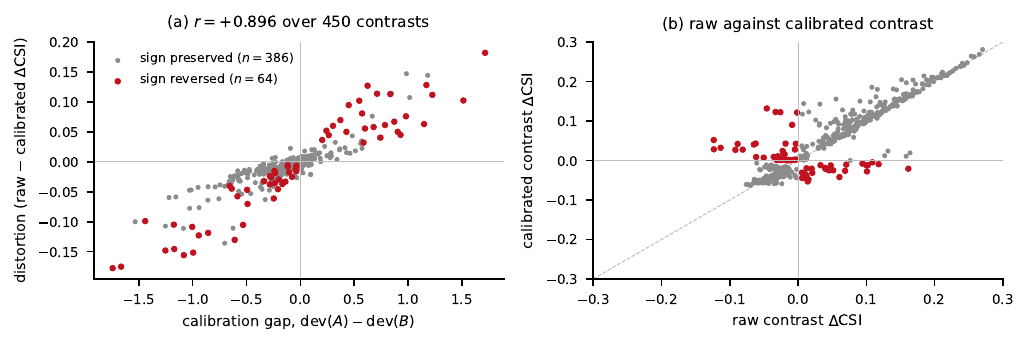}
\caption{All \ExactN\ pairwise contrasts among the \ContrastNArms\ SEVIR arms,
under the exact control of Section~\ref{sec:setup}, which fits the map after
pooling and so cannot reorder pooled cells under any convention.
\textbf{(a)} Distortion (raw minus calibrated CSI contrast) against the pair's
calibration gap. \textbf{(b)} Calibrated against raw contrast; points off the
diagonal moved under the control, and points in the upper-left and lower-right
quadrants reversed sign. Red marks the \ExactFlips\ reversals in both panels.}
\label{fig:contrasts}
\end{figure}

These cells are not independent, so we resample the event, the independent unit,
\CBootN\ times, keeping the contrasts paired within each replicate. The primary
correlation is $\CBootRMax\ [\CBootRMaxLo, \CBootRMaxHi]$ and over all cells
$\CBootRAll\ [\CBootRAllLo, \CBootRAllHi]$. The reversal count is
$\CBootFlips\ [\CBootFlipsLo, \CBootFlipsHi]$ (Appendix~\ref{app:boot}).

\paragraph{Under the exact control.} We re-ran all \ContrastNArms\ arms from
their released checkpoints and repeated every contrast under the exact control
of Section~\ref{sec:setup}. Scored with the pre-pooling global map, the
re-run reproduces the values above ($r = \RerunR$ against \ContrastRAll,
\RerunFlips\ reversals against \ContrastFlips). Under the exact control the
relationship is stronger. Over the \ExactN\ contrasts, $r = \ExactR$ and $\rho =
\ExactRho$, with \ExactFlips\ sign reversals (Figure~\ref{fig:contrasts}), \ExactSep\
of them with both 95\% intervals excluding zero.

Appendix~\ref{app:exact} splits the global-map results by pooling family, which
shows where the two differ. The exact control also corrects the calibrated values at two average-pooled cells. At $16{\times}16$
average pooling and $\tau = 160$ and $181$ the pre-pooling map placed the
CasCast cascade ahead of its backbone (\AvgCascOneSixtyPre\ and
\AvgCascOneEightyOnePre). The exact control places it behind
(\AvgCascOneSixtyExact\ and \AvgCascOneEightyOneExact), which agrees with its
raw contrast under that convention. No raw score is affected, and the
average-pooled reversal count falls from \PaperFamAvgFlips\ to \ExactAvgFlips.

Table~\ref{tab:headline-contrasts} shows, at the most extreme threshold, the
\ContrastNPubPairs\ pairs that published papers themselves compare: CasCast's
cascade against its backbone; the cascade against the official EarthFormer
checkpoint, which \citet{gong2024cascast} evaluate as a baseline for regional
extremes; and EarthFormer against persistence \citep{gao2022earthformer}. The
second is a further audited comparison. In our evaluation the gain falls from \EFCascRaw\ to
\EFCascCal, a \EFCascErased\ relative-gain reduction.
Over the \ContrastNPub\ contrasts of these pairs, $r=\ContrastRPub$ and \ContrastFlipsPub\ reverse sign. The distortion acts in both directions.
Calibration shrinks the cascade's gains over its backbone and over EarthFormer,
but it reveals a larger EarthFormer advantage over persistence. Of all \ContrastN\ contrasts, \ContrastFlips\ reverse sign under the global map and \ExactFlips\ under the exact control, and \ContrastFlipsBig\ of the
former exceed \ContrastFlipsBigThr\ CSI on both sides. The
reversing contrasts are small in absolute CSI but comparable to published
increments in relative terms. Their median raw relative gain is
\ContrastFlipsRelMed, and \ContrastFlipsRelFive\ exceed 5\%. Excluding the
unguided diagnostic arm leaves \ContrastFlipsNoGuid\ reversals. Full threshold-by-threshold results are
in Appendix~\ref{app:boot}, Table~\ref{tab:contrasts}.

\begin{table}[t]
\caption{Our evaluation of the three model pairs compared in prior work, at max $16{\times}16$ pooling and $\tau=219$.
``Raw'' is the relative CSI gain reported without calibration; ``calibrated'' is
the same relative gain after the global held-out quantile map is fitted
separately to both arms. The final column is the reduction in that relative
gain; $\uparrow$ marks a gain that grows. Bias columns give the uncalibrated
pooled frequency bias of each arm.}
\label{tab:headline-contrasts}
\centering
\small
\begin{tabular}{lrrrrr}
\toprule
comparison & bias$_{\rm ref}$ & bias$_{\rm new}$ & raw & calibrated & rel.-gain reduction \\
\midrule
cascade vs. backbone & \EightDetBias & \EightCascBias & \EightReported & \EightResidQ & \EightShareHi \\
cascade vs. EarthFormer & \DeflBiasM & \EightCascBias & \EFCascRaw & \EFCascCal & \EFCascErased \\
EarthFormer vs. persistence & \DeflBiasP & \DeflBiasM & \DeflRaw & \DeflCal & $\uparrow$ \\
\bottomrule
\end{tabular}
\end{table}

\subsection{Two released checkpoints of the same architecture}
\label{sec:samearch}

A more direct check holds architecture fixed. CasCast's deterministic backbone uses
EarthFormer's cuboid-attention architecture, retrained independently on the same
benchmark. At the reference cell the two released checkpoints differ by
\SameArchRaw\ without calibration (Figure~\ref{fig:consequence}a). After both receive the global map the
difference becomes \SameArchCal, with sign reversals at \SameArchFlips\ of
\SameArchN\ thresholds. Their pooled biases are \SameArchBiasA\ and \SameArchBiasB\ (Appendix~\ref{app:samearch}).
The result does not depend on which control is used. Under the exact control of
Section~\ref{sec:setup} the sign reverses at \SameArchExactFlips\ of the
\SameArchExactN\ thresholds. A gap large enough to read as architectural can therefore arise between two trainings of one architecture whose operating-point biases differ. The two groups' recipes differ in more than the seed, so we do not attribute the gap to seed variance. We conclude only that architecture cannot explain it.

\subsection{Independently released comparisons}
\label{sec:independent}

The comparisons above all draw on the same \ContrastNArms\ SEVIR arms. We
therefore repeat the audit on four comparisons between systems released by their
own authors, none of which we trained. Two add SEVIR systems, and two leave the
benchmark (Table~\ref{tab:breadth}).

DiffCast \citep{yu2024diffcast} against its PhyDNet backbone on SEVIR-128 has a
threshold-averaged relative gain at max $16{\times}16$ that falls from
\DCMaxSixteenRawRel\ to \DCMaxSixteenCalRel. At the rarest thresholds the map
removes \DCRemovedLo--\DCRemovedHi\ of the CSI difference, and \DCFlips\ of \DCNLive\
cells reverse sign, \DCFlipsSep\ of them separated (Appendix~\ref{app:exact}).

PreDiff \citep{gao2023prediff} against persistence on SEVIR-LR is the predicted
null case. Both arms sit near unit bias in \PDNearOne\ of \PDNLive\ cells. No
cell whose raw contrast is separated from zero loses more than \PDShareMax\ of
it, and no cell moves by more than \PDMoveMax\ CSI.

Outside SEVIR, on WeatherBench~2 \citep{rasp2024weatherbench} the calibration
gap orders the movement over \WBN\ contrasts ($r = \WBR$), with \WBFlips\
reversals, \WBFlipsSep\ of them separated. At $72$\,h and $50$\,mm, FuXi has frequency
bias \WBFuxiBias. Its CSI difference from IFS~HRES is \WBFuxiRaw\ before
calibration and \WBFuxiCal\ after. On ECMWF operational output the pattern holds over
\ECN\ contrasts ($r = \ECR$), and at the same cell the AIFS advantage over IFS~HRES
in CSI grows from \ECCsiSeventyTwoRaw\ to \ECCsiSeventyTwoCal,
consistent with ECMWF's own report that AIFS under-forecasts heavy precipitation
\citep{moldovan2025update}. The control therefore moves a contrast in the
direction the calibration gap predicts, whether by reversing it or
widening it. Appendix~\ref{app:exact} gives both setups.

\begin{table}[h]
\caption{The number of contrasts in each setting of this section, and how many reverse sign. The bracketed count is the
reversals whose raw and calibrated 95\% intervals both exclude zero.}
\label{tab:breadth}
\centering
\small
\begin{tabular}{llrr}
\toprule
setting & systems & contrasts & reversals \\
\midrule
Six SEVIR systems, exact control & 6 & 450 & 64 (46) \\
DiffCast against PhyDNet, SEVIR-128 & 2 & 30 (28 with events) & 16 (10) \\
WeatherBench 2, 24-h precipitation, 2020 & 5 & 360 & 49 (41) \\
ECMWF operational, 2025--26 & 4 & 270 & 9 (6) \\
PreDiff against persistence, SEVIR-LR & 2 & 30 (29 with events) & 0 (0) \\
 
\bottomrule
\end{tabular}
\end{table}

A line fit on the \ContrastNArms\ SEVIR arms alone predicts the DiffCast and
PreDiff movement out of sample (Appendix~\ref{app:exact}).

\section{Generality beyond one benchmark}
\label{sec:general}

The claim is not specific to radar. On ShanghaiTech Part A we deliberately use
$32{\times}32$ patch sums, a reduction that does not create the SEVIR
max-pooling asymmetry. The identity, a positive gain and a positive shift preserve the order of the patch sums exactly. Image edges are cropped, so every patch is a full $32{\times}32$ block. Predictions are clipped at zero. A shift therefore adds the same $1024\delta$ to every patch sum. Of the \CrowdN\ selected maps, \CrowdOrdN\ are of these kinds. The other \CrowdOrdOther\ can reorder patches. Across trained CSRNet arms the bias deficit still orders the absolute knob gain ($\rho=\CrowdRho$ over \CrowdN\ cells), although the modality, architecture, and pooling rule all differ from SEVIR. The correlation is \CrowdOrdRho\ on the \CrowdOrdN\ cells alone. It is \CrowdPostRho\ with the map fitted after the patch sum.

With the recipe held fixed, seed variation alone produces the effect. Five CSRNet
seeds trained with one recipe span pooled bias
\CrowdSeedBiasLo--\CrowdSeedBiasHi\ and CSI \CrowdSeedCsiLo--\CrowdSeedCsiHi,
and pooled bias orders their CSI at $\rho=\CrowdSeedRho$. Their dense-patch
count errors differ by a factor of \CrowdSeedMaeFold. The much larger CSI
spread is not accompanied by a comparable spread in dense-patch count error
(Appendix~\ref{app:crowdpool}). The identification problem therefore appears across published systems and within one training recipe.

\paragraph{Separating placement from attenuation.} A target-corruption control
separates the two components. Spatial displacement is strongly
penalised by CSI, but leaves pooled bias in
$[\CrowdShiftBiasLo,\CrowdShiftBiasHi]$ and a knob gain of at most
\CrowdShiftGain. A mass-conserving blur likewise remains close to calibration
under patch sums. By contrast, compressing each pixel toward its local mean
creates the deficit (bias \CrowdFlatHardBias) and a knob gain of
\CrowdFlatHardGain. The metric therefore reacts to both placement and amplitude,
and the monotone control measures sensitivity to amplitude adjustment
(full results in Appendix~\ref{app:crowdpool}).

\paragraph{Semantic segmentation as the boundary case.} Per-class intersection
over union has the same contingency-table arithmetic,
\begin{equation}
\mathrm{IoU}=\frac{TP}{TP+FP+FN}=\frac{h}{h+f+m}=\csi.
\end{equation}
We evaluate each class independently at a probability threshold of $0.5$.
This is a binary IoU control rather than standard multiclass argmax evaluation.
We score \SegArch\ on COCO \texttt{val2017} without retraining and apply one held-out monotone logit shift
per class. Here our rarity prediction fails. Over \SegNCls\ classes,
$\rho=\SegRhoRarity$ between event frequency and knob gain, and the mean
movement is only \SegMeanGain\ IoU, because \SegNearOne\ of \SegNCls\ classes
already lie within \SegNearThr\ of $\bias=1$ across a wide range of class
frequencies and leave the knob little deficit to remove.

Where segmentation classes are miscalibrated, the relationship reappears. Knob gain
tracks the bias deficit at $\rho=\SegRhoBias$, rising to \SegRhoBiasPoolA\ under
max $8{\times}8$ pooling. A short-bias class such as \SegBestCls\ gains
\SegBestGain, while an over-forecasting class such as \SegOverCls\ is hurt by
\SegOverGain\ when pulled toward one. Two conditions are therefore needed: a threshold that is fixed in advance, and
a model whose output is poorly calibrated at that threshold. Rarity makes the problem severe for the squared-error-type
regression models we study, but rarity alone is not the diagnostic. Class-level
results and the argmax reproduction check are in Appendix~\ref{app:seg}.

\section{Discussion and recommendations}
\label{sec:disc}

\paragraph{Recommended use.} The practical
recommendation is to report pooled frequency bias beside every rare
pool-and-threshold score and, when compared arms sit at different biases, repeat
the comparison after the same held-out monotone calibration procedure is fitted
separately to each arm, stating the calibration split and the transform
family, using the \texttt{freeknob} package (\CodeURL)
(Appendix~\ref{app:reproducibility} gives the call and how to read what it
returns). We use $\bias=1$ as a common operating point for the comparison. We do not claim that it is the best operating point for any application.

The raw-to-calibrated movement is the part of the contrast that can be generated
without improving spatial ranking, and its share of the contrast is what
we report, because the residual changes sign with the budget. The audit is a control for the
comparison. It does not produce a corrected leaderboard. In the CasCast
decomposition the residual's interval excludes zero under two controls, but
with opposite signs: \EightResidQ\ under the global map
(Appendix~\ref{app:boot}) and \ExactRefCalRel\ under the exact control. The
score therefore cannot show whether the cascade places extremes better than its
backbone. It does show that calibration alone accounts for
\EightShareQ--\ExactRefShare\ of the raw gain.

\paragraph{Amplitude as skill and as confound.} Generative
nowcasters are designed in part to restore the sharp amplitudes that smoother
regression models lose, and getting magnitude right at a physical threshold is
real predictive skill. A pool-and-threshold metric is therefore not at fault
for rewarding amplitude. The identification problem arises when the same
score is read as evidence that an architecture places extremes better. If an
unpaired, order-preserving marginal transform reproduces most of a contrast,
the score alone cannot separate improved placement from improved amplitude.

\paragraph{Scope and limitations.} The relative-gain reductions are measured on
the comparisons we audit, not estimated as a literature-wide average.
Section~\ref{sec:independent} extends the audit to two further released SEVIR
systems and two settings outside SEVIR, so the effect is not specific to one
benchmark, but it remains a finite set of comparisons. Our conclusions
apply only to contingency-table scores at fixed operating points after
pooling. We did not test CRPS, LPIPS, or SSIM. The magnitude of the audit
depends on the admissible transform family as well. Per-cell and per-horizon
controls can leave different residuals (Appendices~\ref{app:lead}
and~\ref{app:knobsevir}).
The audit is a diagnostic. It does not claim that amplitude is unimportant, or
that the residual is purely spatial skill. An order-preserving
calibration map can only account for the part of a difference that such a map
could produce.

\section*{Reproducibility statement}

All reported results are generated programmatically from stored experimental
artefacts rather than entered manually. We evaluate released checkpoints through
a common pipeline, use deterministic calibration/evaluation splits where
applicable, and provide the \texttt{freeknob} implementation of the proposed
audit at \CodeURL. Appendix~\ref{app:reproducibility}
documents the evaluation protocol, statistical conventions, checkpoint settings,
reproduction checks, and model-inclusion decisions needed to reproduce the
experiments.

\section*{AI use statement}

The research question and the identification claim originated with the authors.
An AI assistant helped sharpen that framing and helped design the matched-control
protocol and robustness checks in the appendices. These were then stress-tested
with an AI assistant acting as an adversarial reader. It implemented the
\texttt{freeknob} package and evaluation pipeline, wrote the data-assembly
scripts for the ERA5/Himawari-8 record, and served as a sounding board while we
interpreted intermediate statistics and settled on our readings. It also
drafted and edited prose, produced figures and tables, and helped search for and
summarise related work.

We did not use AI to generate synthetic data, formulate mathematical claims, or
assist with translation. The paper contains no proofs, so proof-related
categories do not apply.

The authors reviewed all AI-assisted work. Code was checked against a reference
implementation that reproduces all thirty SEVIR cells to $10^{-7}$
(Appendix~\ref{app:reproducibility}). Dataset descriptors and references were checked by hand against their sources. AI-drafted prose was revised
throughout. We take responsibility for the final content, including text,
claims, and artefacts produced with AI assistance.

\bibliographystyle{iclr2027_conference}
\bibliography{refs}

\appendix
\raggedbottom

\section{Reproducibility details}
\label{app:reproducibility}

\paragraph{Code, artefacts, and evaluation pipeline.}
Every SEVIR result in this paper is produced from a released checkpoint evaluated
through one common pipeline on the official preprocessed data used in our
experiments. Tables, figures, and the statistics quoted in the text are generated
programmatically from the resulting JSON artefacts. Dataset descriptors, such as
the split dates and base rates of Appendix~\ref{app:atmos}, are transcribed from
the released data audit. The
code repository (\CodeURL) includes the installable \texttt{freeknob} package, whose
\texttt{free\_knob\_audit} function implements the matched-control procedure used
in the paper and returns the corresponding diagnostics and pairwise contrasts:

\begin{center}\vspace{-0.8em}
\fbox{\begin{minipage}{0.84\textwidth}\ttfamily\small
report = free\_knob\_audit(pred[val], obs[val],\\
\hspace*{7.5em}pred[test], obs[test],\\
\hspace*{7.5em}threshold=tau, pooling=pool)
\end{minipage}}\vspace{-0.8em}
\end{center}

\paragraph{Reading and reporting the audit.} Agreement indicates stability under
the specified control. Shrinkage or growth measures its effect on the contrast. A sign reversal prevents reading the raw ordering as an ordering of spatial
discrimination. None of these readings treats $\bias=1$ as application-optimal.
Near-unit pooled bias usually accompanies small movement. Exceptions warrant
checking the fitted map and the raw-versus-pooled marginal distinction
(Appendices~\ref{app:seg} and~\ref{app:valfit}).

\paragraph{Empirical quantile-map conventions.}
The implementation linearly interpolates paired forecast and observed quantile
knots and extrapolates using the endpoint segments. A left-sided search selects
the segment at a knot. A repeated source knot uses the lower endpoint when the
segment has zero width. The same deterministic rule is applied to equal inputs.
These conventions resolve empirical ties without introducing pixel-specific information.

The supplied parity record covers all thirty SEVIR pooling--threshold cells for
the CasCast deterministic backbone and agrees with the reference computation to
an absolute tolerance of $10^{-7}$. The repository's verification script checks this stored
record. Re-running the complete parity harness additionally requires the external
data and prediction files.

\paragraph{Splits and statistical conventions.}
For the main SEVIR experiments, calibration and evaluation events are separated
by index parity, so the split requires no random seed. Unless stated otherwise,
rank correlations are Spearman's $\rho$ with tied ranks assigned their average
rank, and the associated $p$-values are computed by exact permutation tests.
This convention matters because several evaluated quantities contain exact ties.

\paragraph{SEVIR and CasCast settings.}
Predictions are handled in VIL units using the released repository's $x/255$
convention. CasCast's diffusion cascade is sampled with classifier-free guidance
weight $2$ and one ensemble member, i.e.\ one sample per event, matching the
authors' released evaluation script. These choices materially affect the
extreme-threshold results. Our re-run of the released deterministic backbone
agrees with the published cells to within \ReproWorst. The cell-level comparison
is reported below.

\subsection{Reproduction of the CasCast backbone}
\label{app:repro}

Table~\ref{tab:repro} compares our re-run of the released deterministic
checkpoint with the four cells reported by \citet{gong2024cascast}. This
reproduction check is what allows the analysis in
Section~\ref{sec:cascast} to study a released model rather than a
re-implementation.

\begin{table}[h]
\caption{Reproduction of the CasCast deterministic backbone against the values
published in the authors' Table~4. CSI-M is the mean over the six thresholds.}
\label{tab:repro}
\centering
\begin{tabular}{llrrr}
\toprule
metric & pooling & published & ours & deviation \\
\midrule
CSI-M & none & 0.4310 & 0.4299 & -0.2\% \\
CSI-M & max $16^2$ & 0.4351 & 0.4327 & -0.5\% \\
CSI-219 & none & 0.1448 & 0.1488 & +2.8\% \\
CSI-219 & max $16^2$ & 0.1481 & 0.1467 & -0.9\% \\
 
\bottomrule
\end{tabular}
\end{table}

\paragraph{Model inclusion and exclusion.}
We restrict the primary comparison to systems for which the relevant released
artefacts can be evaluated faithfully. DGMR \citep{ravuri2021dgmr} and
NowcastNet \citep{zhang2023nowcastnet} provide released checkpoints, but those
checkpoints were trained on different radar products, resolutions, and temporal
sampling from SEVIR. Evaluating either on SEVIR would therefore require
retraining, turning the comparison into one against our re-implementation rather
than the released model.

PreDiff \citep{gao2023prediff} is evaluated in Section~\ref{sec:independent}.
Its original released weight URLs no longer resolve. We recovered the weights
from the authors' repository and scored $2{,}027$ test windows with one sample
each. Our pooled CSI is within $3\%$ of the published value, and $12\%$ below it
at pixel level, a discrepancy we have not resolved and report as such. We
include the arm because the audit's conclusion for it does not turn on that
offset. Both arms sit near unit frequency bias, so the control moves the contrast
very little either way.

The two exclusions affect the breadth of the empirical comparison but not the
definition of the audit. Any additional checkpoint can be evaluated with the
same pooled-frequency-bias diagnostic and symmetric held-out calibration
procedure.

\section{Per-class segmentation results}
\label{app:seg}

Table~\ref{tab:seg} gives the per-class results behind
Section~\ref{sec:general}, using DeepLabv3 \citep{chen2017deeplabv3}
on COCO \citep{lin2014coco}. Every class with enough calibration support to fit a
knob is shown, ordered by positive-pixel fraction. The gain column does not
follow that ordering.

One class shows that near-unit bias does not guarantee an inert knob.
\SegCtrCls\ sits at bias \SegCtrBias, inside the
\SegNearThr\ band, and still loses \SegCtrGain\ IoU. It is a \SegCtrR\ class, so
the bias estimated on the calibration half is noisy, and the shift that zeroes it
there is the wrong shift on the evaluation half. Re-selecting the same family on
validation IoU rather than on $|\bias - 1|$ recovers most of it,
\SegCtrGainVal\ instead of \SegCtrGain. Half-to-half fitting noise is one possible explanation for the residual harm at
low support, but this comparison does not establish it as the unique cause.

Across the \SegNearOne\ classes inside the near-unit-bias band,
\SegNearFlat\ move by less than \SegFlatThr\ IoU, \SegNearHurt\ lose,
\SegNearHelped\ gain, and the mean change is \SegNearMeanGain. Thus the
average movement is small in this band, although individual classes can move
appreciably.

Our mIoU is \SegReproDev\ relative to the published figure. The gap reflects
two protocol differences: (i)~\SegDiffA; and (ii)~\SegDiffB. We keep the first
because restricting to that subset would select images on the classes we are
measuring, and the second because deleting those regions would remove exactly the
crowded scenes where a rare class is hardest to place. Both choices lower our
mIoU relative to the published number, and neither affects the comparison made
here, which is each class against itself before and after one monotone shift.

\begin{table}[h]
\caption{COCO \texttt{val2017}, \SegArch, released checkpoint. Each class is
evaluated independently as a binary mask at probability threshold $0.5$;
per-class IoU $\equiv\csi$ is reported at that operating point and after one
monotone logit shift $\delta$ fitted on the disjoint calibration half by
$|\bias - 1|$. This audit is distinct from the separate multiclass argmax
reproduction check. Classes are ordered by positive-pixel fraction; those
without enough calibration support to fit a knob are omitted rather than
fitted on noise.}
\label{tab:seg}
\centering
\small\setlength{\tabcolsep}{4pt}
\begin{tabular}{lrrrr@{\hskip 1.2em}lrrrr}
\toprule
class & $R$ & $\bias$ & IoU & $\Delta$ & class & $R$ & $\bias$ & IoU & $\Delta$ \\
\midrule
person & 9.012\% & 1.022 & 0.8155 & -0.0008 & tvmonitor & 0.414\% & 0.963 & 0.6992 & -0.0248 \\
diningtable & 3.002\% & 1.732 & 0.3802 & -0.0351 & horse & 0.336\% & 0.997 & 0.6969 & +0.0010 \\
bus & 0.868\% & 1.005 & 0.7802 & -0.0123 & cow & 0.299\% & 0.964 & 0.7718 & +0.0000 \\
train & 0.788\% & 0.975 & 0.7764 & +0.0000 & sheep & 0.258\% & 0.988 & 0.7033 & -0.0143 \\
chair & 0.729\% & 0.643 & 0.3384 & +0.0204 & aeroplane & 0.227\% & 0.999 & 0.8153 & +0.0000 \\
sofa & 0.702\% & 0.825 & 0.5066 & +0.0226 & bottle & 0.216\% & 0.780 & 0.5125 & +0.0207 \\
dog & 0.615\% & 1.003 & 0.7203 & +0.0000 & bicycle & 0.201\% & 0.942 & 0.6093 & -0.0019 \\
cat & 0.583\% & 0.945 & 0.8360 & +0.0000 & boat & 0.171\% & 0.771 & 0.4487 & +0.0272 \\
motorbike & 0.552\% & 0.899 & 0.7270 & +0.0051 & bird & 0.163\% & 0.861 & 0.6220 & -0.0172 \\
car & 0.551\% & 0.846 & 0.5242 & +0.0000 & pottedplant & 0.156\% & 1.093 & 0.3490 & -0.0311 \\
 
\bottomrule
\end{tabular}
\end{table}

\section{Full results of the rarity sweep}
\label{app:rarity}

This appendix gives the per-threshold results behind Section~\ref{sec:rarity}.
The model is an attention U-Net trained with a weighted mean squared error on
geostationary infrared band B08, with the event defined as $y \le \tau$ kelvin,
so that rarity rises down the table. The model is retrained at every threshold, so every row is scored at the
operating point its own model was trained for. The fraction $R$ and the physical intensity of
the event still move together down the table, as they must on real fields. Each row averages \EoneNRuns\ runs,
five seeds crossed with the two single-decoder loss variants,
\emph{\EoneVariants}. Graded weights each pixel continuously by how far it lies
past the threshold. Tiered places hard breakpoints on the target and weights the
resulting bands. Averaging over a factor can manufacture an ordering, so the
sweep is also computed inside each variant. It gives $\rho = \EoneRhoGraded$ for graded
and $\rho = \EoneRhoTiered$ for tiered against $\log_{10} R$, against
$\EoneRho$ pooled, with exact two-sided permutation $p$ of \EonePGraded,
\EonePTiered\ and \EoneP\ respectively. We attach no significance to the pooled
correlation being the strongest of the three. The test has limited resolution. Six levels
admit $6! = 720$ orderings, so the smallest two-sided $p$ available is \EoneP\
and only a near-perfect ordering reaches conventional significance. The tiered
variant does not. Every departure from a perfect ordering falls at the three most
common thresholds. There the gain is within a few percent of zero in both
variants, and its sign is set by out-of-sample fitting noise rather than by
rarity. Across the three rarest thresholds, where the claim is made, both
variants are separately monotone. The absolute and median columns of
Table~\ref{tab:rarity} support the claim without a rank test.
Here $R$ is the all-scene fraction on the valid evaluation footprint, while the
coverage-filtered score excludes scenes with observed coverage at or below
0.1\%.
Because the retained cohort shrinks with $\tau$, we separately recompute pooled
contingency scores over all 150 scenes in Appendix~\ref{app:atmos}.

\begin{table}[h]
\caption{Gain from a free monotone recalibration against event rarity. $R$ is
the all-scene test fraction on the valid evaluation footprint; scores retain
the observed-coverage $>0.1\%$ frame filter. The
relative gain is undefined for a run whose uncalibrated $\csi$ is zero, so $n$
gives the runs entering that column against the \EoneNRuns\ trained at that
$\tau$. Columns three to six are means over those $n$ runs, which is why
$\Delta\csi$ and $\Delta\csi$ (all) differ at the two rarest levels; the last two
columns are computed over all \EoneNRuns. The \EoneExcluded\ runs the relative
column omits score exactly zero uncalibrated and are lifted to between
\EoneExclLo\ and \EoneExclHi\ $\csi$ by the knob, so their true relative gain is
unbounded and dropping them can only understate the trend.}
\label{tab:rarity}
\centering
\small\setlength{\tabcolsep}{4.5pt}
\begin{tabular}{rrrrrrrrr}
\toprule
$\tau$ & $R$ (test) & $\csi$ identity & $\csi$ recal & $\Delta\csi$ &
rel.\ gain & $n$ & $\Delta\csi$ (all) & median (all) \\
\midrule
220 & 20.270\% & 0.2672 & 0.2610 & -0.0062 & -2.3\% & 10/10 & -0.0062 & -0.0048 \\
215 & 11.537\% & 0.1992 & 0.1956 & -0.0036 & -1.8\% & 10/10 & -0.0036 & -0.0011 \\
210 & 6.378\% & 0.1381 & 0.1368 & -0.0013 & -0.9\% & 10/10 & -0.0013 & -0.0039 \\
205 & 3.124\% & 0.0578 & 0.0739 & +0.0160 & +27.7\% & 10/10 & +0.0160 & +0.0143 \\
200 & 0.884\% & 0.0117 & 0.0352 & +0.0236 & +202.1\% & 9/10 & +0.0238 & +0.0245 \\
197 & 0.264\% & 0.0085 & 0.0299 & +0.0214 & +250.8\% & 7/10 & +0.0237 & +0.0265 \\
 
\bottomrule
\end{tabular}
\end{table}

\section{The infrared downscaling task}
\label{app:atmos}

\paragraph{Data and split.}
We construct this setting rather than adopting a packaged benchmark. A
six-hourly scene enters the record when the Guan--Waliser global catalogue
\citep{guan2015atmospheric} marks any atmospheric-river object inside
$20$--$27^\circ$N, $88$--$93^\circ$E. The usable intersection with
geostationary observations contains 1,500 scenes from 2015-07-17 06:00 through
2023-12-17 06:00. The target is Himawari B08 brightness temperature
\citep{bessho2016himawari} on a $0.02^\circ$ grid. The stored ERA5
\citep{hersbach2020era5} predictors are integrated-vapour-transport
magnitude, temperature at 500 and 850 hPa, relative humidity at 700 hPa, and
vertical velocity at 500 hPa, linearly interpolated to that grid. The scored
models use the latter four, with integrated vapour transport stored but
excluded. Predictor means and population standard deviations are fitted per
channel on training pixels only.

We split selected scenes chronologically: 1,200 train
(2015-07-17 06:00--2021-11-14 18:00), 150 validation
(2021-11-15 06:00--2023-03-19 06:00), and 150 test
(2023-03-19 12:00--2023-12-17 06:00). This avoids random-split leakage between
nearby selected times, but it creates temporal shift. At 220 K the full-grid
positive fractions are 18.75\%, 13.95\%, and 19.62\%, respectively. Thus the
validation window is the quietest, and the operating point selected there is
applied across that shift. Moreover, six-hourly scenes are autocorrelated, so
the effective sample size is below the nominal scene count. We disclose both as
limitations and report the no-frame-filter pooled sensitivity below.

\paragraph{Geometry, cohort and base rate.}
Stored targets are $351\times251$. Evaluation follows the training pipeline's
centre crop to a valid $256\times251$ footprint inside its nominal
$256\times256$ tensor. In the tables, $R$ always means the fraction of valid
test pixels with $y\leq\tau$ over all 150 test scenes on that
footprint. At 220 K this is 20.270\%, against 19.622\% on the full grid. The
coverage-filtered CSI protocol then scores only scenes whose observed coverage
exceeds 0.1\%. The event-conditioned training climatology on this footprint is
26.007\% at 220 K. These quantities are intentionally different and are emitted
together by the released base-rate audit.

As a sensitivity check, we also retain every validation and test scene and form
one pooled contingency table per run. The rarity ordering becomes less exact
($\rho=\AtmosAllRho$, $p=\AtmosAllP$), but the knob gain still grows
from \AtmosAllGainCommon{} at 220 K to \AtmosAllGainRarest{} at 197 K. The
rarity association therefore does not depend on the threshold-dependent frame
cohort.

\section{Full results by pooling convention}
\label{app:pooling}

Table~\ref{tab:pooling} gives the detail behind Section~\ref{sec:pooling}. It shows one arm
at one threshold under every pooling convention, with the ranking of the 2019
advection scheme against the 2022 transformer counted beside it.

\begin{table}[h]
\caption{Pooling convention determines the size of the calibration effect and
the ranking. Bias and CSI are for the CasCast deterministic backbone at
$\tau=219$. The last three columns count, out of six thresholds, how often
pysteps (2019 advection) outranks EarthFormer (2022) when both are left
uncalibrated, both receive the global quantile map, and both receive the
per-cell knob.}
\label{tab:pooling}
\centering
\begin{tabular}{lrrrrrrr}
\toprule
\multirow{2}{*}{pooling} & \multirow{2}{*}{$\bias$} &
\multirow{2}{*}{$\csi$} & \multirow{2}{*}{$\csi{+}$qmap} &
\multirow{2}{*}{gain} & \multicolumn{3}{c}{pysteps $>$ EarthFormer} \\
\cmidrule(lr){6-8}
 & & & & & uncal. & +global & +per-cell \\
\midrule
none & 0.291 & 0.1488 & 0.2074 & +39.3\% & 0 & 0 & 0 \\
avg $4^2$ & 0.350 & 0.1724 & 0.2185 & +26.7\% & 0 & 0 & 0 \\
max $4^2$ & 0.214 & 0.1426 & 0.2297 & +61.1\% & 0 & 0 & 0 \\
avg $16^2$ & 0.664 & 0.2347 & 0.2138 & -8.9\% & 0 & 0 & 0 \\
max $16^2$ & 0.172 & 0.1467 & 0.2639 & +79.8\% & 5 & 5 & 0 \\
 
\bottomrule
\end{tabular}
\end{table}

\section{Full crowd-density results under the patch-sum protocol}
\label{app:crowdpool}

This appendix gives the CSRNet \citep{li2018csrnet} results behind Section~\ref{sec:general}. The pool-and-threshold score is our construction. ShanghaiTech \citep{zhang2016shanghaitech} itself reports MAE and RMSE on image counts. The domain contributes a sum reduction, the convention that does not induce a deficit on SEVIR. Table~\ref{tab:crowdpool} gives the trained arms and the constructed controls together, and Table~\ref{tab:crowdseed} the per-seed spread behind the same section.

Section~\ref{sec:general} quotes the bias--gain rank correlation over the
trained arms alone, $\rho = \CrowdRho$ over \CrowdN\ cells, because those are the
analogue of the SEVIR statistic, which ranges over arms. Over all
\CrowdNAll\ cells, controls included, it reads \CrowdRhoAll. We do not report
that figure in the main text. The controls make up most of the cells and form a
large tie at exactly zero gain, so the statistic depends on how that tie is
ranked.

\begin{table}[h]
\caption{Crowd density under the patch-sum protocol of Section~\ref{sec:general}
($32{\times}32$ patch sums), averaged over \CrowdSeeds\ seeds. $\Delta\csi$ is the absolute change from
one monotone knob, selected on a held-out split by $|\bias - 1|$ and never on
$\csi$. The last two rows are targets we corrupted deliberately, to move
placement and sharpness one at a time. Selecting on $|\bias - 1|$ cannot tune
for the score we report, but it is not neutral toward an arm whose bias is
long; re-selecting the same family on validation $\csi$ removes the harm
and leaves every conclusion unchanged (Appendix~\ref{app:knobrule}).}
\label{tab:crowdpool}
\centering
\small
\begin{tabular}{lrrrrrr}
\toprule
\multirow{2}{*}{arm} & \multicolumn{3}{c}{pooled $\bias$} &
                       \multicolumn{3}{c}{$\Delta\csi$ from the knob} \\
\cmidrule(lr){2-4}\cmidrule(lr){5-7}
 & $\tau{=}3$ & $\tau{=}7$ & $\tau{=}10$
 & $\tau{=}3$ & $\tau{=}7$ & $\tau{=}10$ \\
\midrule
CSRNet & 1.122 & 0.682 & 0.289 & +0.0072 & +0.0352 & +0.0911 \\
CNN baseline (MSE) & 1.000 & 1.128 & 1.426 & +0.0021 & -0.0339 & -0.0442 \\
target, displaced 2 blocks & 0.938 & 0.954 & 0.969 & -0.0009 & +0.0000 & +0.0000 \\
target, blurred (mass conserved) & 0.995 & 0.975 & 0.976 & +0.0000 & +0.0000 & +0.0000 \\
target, peaks compressed (mass conserved) & 0.942 & 0.831 & 0.761 & +0.0047 & +0.0000 & +0.0809 \\
 
\bottomrule
\end{tabular}
\end{table}

Table~\ref{tab:crowdseed} is the seed spread quoted in
Section~\ref{sec:general}, with a count-level error beside the score. The seed at $\bias = \CrowdSeedBiasLo$ has an error on the dense patches of
\CrowdSeedMaeHi\ persons against \CrowdSeedMaeLo\ for the best seed, a factor
of \CrowdSeedMaeFold, while its $\csi$ is \CrowdSeedFold\ times smaller. Thus,
the much larger CSI spread is not accompanied by a comparable spread in
dense-patch count error. This comparison is consistent with the attenuation
mechanism of Section~\ref{sec:mechanism}, but does not by itself establish that
every training run converged adequately. The patch error orders the score perfectly across the seeds, $\rho = \CrowdSeedMaeRho$, whereas pooled bias orders it at $\rho = \CrowdSeedRho$ with a single adjacent inversion. This does not show that the score tracks placement. Patch MAE on the dense patches is itself amplitude-sensitive. A flattened density map under-counts where the count is large, so patch MAE and pooled bias both measure the same attenuation, the former at finer resolution. In addition, a
patch counts as positive only if it holds at least
$\tau = \CrowdSeedTau$ people, so a predictor that output zero everywhere would
score a patch error of at least \CrowdSeedTau\ on exactly these patches. The
worst seed's \CrowdSeedMaeHi\ is below that bound, and the best seed's
\CrowdSeedMaeLo\ well below. Every seed performs better than the corresponding all-zero-predictor bound on
this dense-patch error, while the score separates them \CrowdSeedFold-fold. We report patch-level error because the artefacts retain it per cell. Image-level count MAE, this benchmark's principal metric, was not cached per seed and we do not reconstruct it.

\begin{table}[h]
\caption{CSRNet on ShanghaiTech Part A, \CrowdSeeds\ seeds, one recipe, at
$\tau = \CrowdSeedTau$ under the patch-sum protocol. ``patch MAE'' is the mean
absolute count error over patches above the threshold and ``bg MAE'' over the
rest. Only the seed changes across rows.}
\label{tab:crowdseed}
\centering
\small
\begin{tabular}{lrrrr}
\toprule
seed & pooled $\bias$ & $\csi$ & patch MAE & bg MAE \\
\midrule
s42 & 0.002 & 0.0017 & 8.12 & 0.794 \\
s43 & 0.107 & 0.0777 & 6.23 & 0.806 \\
s44 & 0.136 & 0.0687 & 6.38 & 0.885 \\
s45 & 1.146 & 0.3194 & 4.72 & 0.798 \\
s46 & 0.053 & 0.0368 & 6.61 & 1.017 \\
 
\bottomrule
\end{tabular}
\end{table}

\section{Two released checkpoints of one architecture}
\label{app:samearch}

Both arms of Table~\ref{tab:samearchapp} are the same cuboid transformer.
CasCast's deterministic backbone reuses EarthFormer's attention code, in the
file \texttt{networks/earthformer\_xy.py}, and was trained by the CasCast authors.
This is not a
published head-to-head comparison, and we do not present it as one. It measures
how large an apparent architectural gap can arise between two trainings of one
architecture.

\begin{table}[h]
\caption{Two independent trainings of one architecture, max $16{\times}16$
pooling. ``raw'' columns are what a results table would print; ``calibrated''
columns follow one global quantile map fitted per arm on the calibration half.
The sign of the contrast reverses at \SameArchFlips\ of the \SameArchN\
thresholds.}
\label{tab:samearchapp}
\centering
\begin{tabular}{rrrrrrr}
\toprule
$\tau$ & bias EF & bias CasCast & raw $\Delta\csi$ & cal.\ $\Delta\csi$ &
raw & calibrated \\
\midrule
16 & 0.915 & 0.875 & -0.0109 & +0.0065 & -1.4\% & +0.9\% \\
74 & 0.747 & 0.718 & -0.0149 & -0.0007 & -2.2\% & -0.1\% \\
133 & 0.534 & 0.495 & -0.0236 & +0.0055 & -5.1\% & +1.1\% \\
160 & 0.387 & 0.337 & -0.0315 & +0.0054 & -9.5\% & +1.4\% \\
181 & 0.337 & 0.285 & -0.0324 & +0.0118 & -11.3\% & +3.4\% \\
219 & 0.281 & 0.172 & -0.0614 & +0.0133 & -29.5\% & +5.3\% \\
 
\bottomrule
\end{tabular}
\end{table}

\section{Fitting the transform on an earlier validation window}
\label{app:valfit}

Every knob in the main text is fitted on the calibration half of the test split
and evaluated on the other half. That control addresses the paper's question,
whether a comparison between two arms survives giving both the same free
transform, and a symmetric split is appropriate for it. It leaves open whether
the transform remains effective when it must be fitted in advance and applied
across temporal drift.

SEVIR's preprocessed nowcast release ships a training file and a testing file and
no validation file, so the split has to be reconstructed. EarthFormer's loader
takes \texttt{train\_val\_split\_date} $=$ 2019-01-01 and CasCast's file lists
name their validation events \texttt{2019\_0101\_0430}, so both hold out the same
window. The test file begins on 2019-06-11. That window is therefore held out
from both models we decompose and lies entirely before the evaluation period. We
run all six arms over \ValN\ events sampled evenly across it, fit one global
quantile map per arm there, and apply it unchanged to the same test evaluation
half the rest of the paper scores. Table~\ref{tab:valfit} gives the result at the
reference cell.

\begin{table}[h]
\caption{The same arms at max $16{\times}16$ pooling and $\tau = 219$, scored on
the test evaluation half, with the global quantile map fitted two ways.
``test-fit'' is the paper's default, on the disjoint calibration half of the test
split; ``val-fit'' is on \ValN\ events from 2019-01-01 to 2019-04-30, which
precede the test period and were held out of both models' training. The
backbone's val-fit score comes from the earlier inference run described in
Table~\ref{tab:boot}.}
\label{tab:valfit}
\centering
\small\setlength{\tabcolsep}{4.5pt}
\begin{tabular}{lrrrrrr}
\toprule
arm & $\bias$ & $\csi$ & $\csi$ test-fit & $\csi$ val-fit & gain test-fit &
gain val-fit \\
\midrule
persistence & 1.020 & 0.1924 & 0.1908 & 0.1936 & -0.8\% & +0.6\% \\
pysteps & 0.907 & 0.2700 & 0.2704 & 0.2719 & +0.2\% & +0.7\% \\
EarthFormer & 0.281 & 0.2082 & 0.2505 & 0.2877 & +20.3\% & +38.2\% \\
CasCast backbone & 0.172 & 0.1467 & 0.2639 & 0.2928 & +79.8\% & +99.6\% \\
CasCast cascade & 1.037 & 0.3069 & 0.2978 & 0.3063 & -2.9\% & -0.2\% \\
CasCast cascade (no guidance) & 0.548 & 0.2475 & 0.2899 & 0.2662 & +17.1\% & +7.6\% \\
 
\bottomrule
\end{tabular}
\end{table}

The conclusion does not depend on which fit is used. The main decomposition reports
\ValHeadRaw\ raw and retains \ValHeadTest\ under the test-half fit against
\ValHeadVal\ under the val-fit, so the relative-gain reduction is
\ValErasedTest\ against \ValErasedVal. Over all \ValCells\ pairwise contrasts the calibration gap
is associated with the distortion at $r = \ValRTest$ under the test-half fit
and $r = \ValRVal$ under the val-fit, with \ValFlipsTest\ and \ValFlipsVal\ sign
reversals respectively.

Fitting out of period does not reduce the knob gain. Across the
\ValNPay\ arms that gain from the knob, the val-fitted map is worth between
\ValRatioLo\ and \ValRatioHi\ of what the test-half fit is worth, least for
\ValRatioLoArm\ and most for \ValRatioHiArm. On \ValFlatArms\ it moves CSI by
at most \ValFlatMax\ under either fit, which is the persistence control of
Section~\ref{sec:mechanism} repeated with a different fitting set. Those two are
named rather than described as the arms near $\bias = 1$, because pooled bias is
not what selects them. The cascade sits at pooled bias \ValCascBias, closer to one
than pysteps at \ValFlatOtherBias\ by a factor of \ValCascCloser\ in
$|\log \bias|$. The knob still leaves pysteps unchanged while moving the cascade
\ValCascTest\ test-fit and \ValCascVal\ val-fit. As noted in
Section~\ref{sec:setup}, the map is fitted on the raw, unpooled marginal while
the metric is computed after pooling, so a small pooled bias deficit does not by
itself imply an inert knob. We do not interpret the ratios above one as evidence that an earlier,
smaller fitting sample is better. A single window cannot support that. The
claim they support is the narrow one, that the knob gain is not an artefact of
fitting inside the split it is scored on.

The val-fit column of Table~\ref{tab:valfit} is a point estimate. The interval we quote on the residual, in
Appendix~\ref{app:boot}, is fitted on the test calibration half, which is the
control the paper's claim is stated against. We attach no interval to the val-fit
residual and rest nothing on its significance.

\section{The guidance arm}
\label{app:guid}

Classifier-free guidance weight is a sampling-time scalar that is set after
training and changes no model parameter. On this benchmark it nonetheless moves
pooled frequency bias. At max $16{\times}16$ and $\tau = 219$ the ladder runs
backbone, unguided cascade, guided cascade: pooled bias \EightDetBias,
\GuidBias, \EightCascBias, and $\csi$ \EightDet, \GuidCsi, \EightCasc. The free
knob gain falls along the same ladder, \EightDetGain, \GuidGain,
\EightCascGain, which is the bias--gain relationship of Section~\ref{sec:mechanism} inside
one system's own decoding settings. The ladder splits into two legs. Diffusion
without guidance covers \LegDiffLog\ of the backbone-to-guided distance in
$|\log \bias|$ and \LegDiffLin\ of it on bias directly. The guidance weight
covers the remaining \LegCfgLog\ and \LegCfgLin\ respectively. We quote
$|\log \bias|$ first because it is the deviation the pairwise contrasts of
Section~\ref{sec:contrasts} are ordered by. The two scales disagree about
which leg is larger, so we state the measure explicitly. A comparison that reports the
cascade against the backbone at a fixed guidance weight is reporting a decoding
choice alongside an architectural one, and on the measure this paper uses
throughout the decoding choice is the smaller of the two. More generally, at a
rare threshold under max pooling, any change that restores amplitude is scored
as though it had improved placement.

\section{Selection rules for the transform}
\label{app:knobrule}

The measured knob gain depends on both the transform family and its selection
rule. A richer family or selection by validation CSI can change either arm's
score, but need not reduce the between-arm residual or improve evaluation CSI. It is also not neutral between arms.
Selecting on
$|\bias - 1|$ cannot tune for the score we report, but it is not neutral toward
an arm whose bias is long. Pulling such an arm to $\bias = 1$ removes
hits faster than false alarms at a rare threshold, and costs it $\csi$. The MSE
baseline in Table~\ref{tab:crowdpool} is that case, and under this rule the knob
takes as much as \CrowdBaseBiasMin\ $\csi$ from it. Re-selecting the same family
to maximise validation $\csi$, which gives each arm its best shot while
still never fitting on test, consumes paired labels rather than the observed
marginal alone, so it is outside the ``unpaired statistics only'' description the
rest of the paper claims and is reported as a robustness check rather than as the
control. It all but removes the harm (worst case \CrowdBaseValMin, from \CrowdBaseBiasMin) and leaves the
conclusion unchanged. CSRNet still recovers up to \CrowdCsrValMax, more than the
\CrowdCsrBiasMax\ the conservative rule grants it. We report the conservative
rule throughout for that reason.

On SEVIR the two rules differ even less. At the reference cell the two rules select the same member,
\KnobDetKnob, so the backbone reaches \KnobDetPooledBias\ either way. Across the
cells of Appendix~\ref{app:knobsevir} the validation-CSI rule is not uniformly
better. At \KnobCsiLosePool, $\tau = \KnobCsiLoseTau$ it selects a member scoring
\KnobCsiLoseCsi\ on the evaluation half against the bias rule's
\KnobCsiLoseBias, having fitted the calibration half's noise. Selecting on the
score can lose out of sample what it gains in sample, which is a second reason to
report the rule that cannot tune for the score at all.

\section{Decomposition under the equitable threat score}
\label{app:ets}

Table~\ref{tab:ets} is the decomposition of Section~\ref{sec:cascast} recomputed
under the equitable threat score, $\mathrm{ETS} = (h - h_r)/(h + f + m - h_r)$
with $h_r = (h+f)(h+m)/N$ and $N$ the number of pooled blocks over the
evaluation half. It reuses the per-event count cache, so the two columns are the
same hits, false alarms and misses read by two scores rather than two
experiments.

Chance correction does not remove the confound. Under ETS the calibration
transform erases between \EtsErasedLo\ and \EtsErasedHi\ of the cascade's gain
across the \EtsNThr\ thresholds at which there is a gain to erase, and at no
threshold does the relative-gain reduction differ from the $\csi$ figure by
more than \EtsMaxGap\ points. The share is left blank at $\tau = \EtsThrNoGain$, where the cascade
does not lead the backbone in the first place and a fraction of the gain is not
defined. That row is undefined under both scores. This result is expected. The chance term $h_r$ is computed at the forecast's
own frequency, so a sharper forecast is charged a larger chance term, but the
charge is calibrated to a random forecast and not to the geometry by which
pooling converts sharpness into hits. Equitability is a statement about
$\mathbb{E}[\text{score}]$ under a random forecast. It is not a statement about
bias sensitivity for skilled ones \citep{hamill2006climatology}.

\begin{table}[h]
\caption{The cascade-over-backbone gain of Section~\ref{sec:cascast} at max
$16{\times}16$, raw and after both arms receive the same monotone knob, scored
under $\csi$ and under ETS. ``rel.-gain reduction'' is the fraction of the raw relative gain removed by the knob.}
\label{tab:ets}
\centering
\small
\begin{tabular}{rrrrrrr}
\toprule
& \multicolumn{3}{c}{$\csi$} & \multicolumn{3}{c}{ETS} \\
\cmidrule(lr){2-4}\cmidrule(lr){5-7}
$\tau$ & raw & calibrated & rel.-gain reduction &
raw & calibrated & rel.-gain reduction \\
\midrule
16 & -0.2\% & +4.9\% & --- & -1.8\% & +5.0\% & --- \\
74 & +7.4\% & +4.7\% & 36\% & +7.4\% & +4.4\% & 41\% \\
133 & +26.9\% & +12.2\% & 55\% & +27.0\% & +11.9\% & 56\% \\
160 & +52.4\% & +14.8\% & 72\% & +52.0\% & +14.3\% & 73\% \\
181 & +64.7\% & +14.5\% & 78\% & +64.1\% & +14.0\% & 78\% \\
219 & +109.1\% & +12.9\% & 88\% & +108.2\% & +12.5\% & 88\% \\
 
\bottomrule
\end{tabular}
\end{table}

\section{Bootstrap intervals}
\label{app:boot}

\begin{table}[t]
\caption{Full threshold-by-threshold version of the three contrasts summarised in the main text, at max
$16{\times}16$ pooling. The first block runs from $\tau=74$ because the
relative-gain reduction quoted in the text starts there; the others start at
$133$. ``raw'' is the
relative gain as a results table would
print it, ``calibrated'' the same gain after one global quantile map is fitted
per arm on the calibration half and applied to both, and
``rel.-gain reduction'' is
$1 - \text{calibrated}/\text{raw}$, with $\uparrow$ marking rows where
calibration raises the gain so that ratio has no stable base. In each
label ``X vs.\ Y'', Y is the reference arm and X the one claimed better, and the
bias columns are their uncalibrated pooled biases in that order.}
\label{tab:contrasts}
\centering
\small\setlength{\tabcolsep}{4.5pt}
\begin{tabular}{lrrrrrr}
\toprule
comparison & $\tau$ & bias$_{\text{ref}}$ & bias$_{\text{new}}$ & raw &
calibrated & rel.-gain reduction \\
\midrule
\multirow{5}{*}{cascade vs.\ own backbone} & 74 & 0.718 & 0.961 & +7.4\% & +4.7\% & 36\% \\
 & 133 & 0.495 & 0.905 & +26.9\% & +12.2\% & 55\% \\
 & 160 & 0.337 & 0.875 & +52.4\% & +14.8\% & 72\% \\
 & 181 & 0.285 & 0.904 & +64.7\% & +14.5\% & 78\% \\
 & 219 & 0.172 & 1.037 & +109.1\% & +12.9\% & 88\% \\
\midrule
\multirow{4}{*}{cascade vs.\ EarthFormer} & 133 & 0.534 & 0.905 & +20.5\% & +13.5\% & 34\% \\
 & 160 & 0.387 & 0.875 & +38.0\% & +16.4\% & 57\% \\
 & 181 & 0.337 & 0.904 & +46.0\% & +18.3\% & 60\% \\
 & 219 & 0.281 & 1.037 & +47.4\% & +18.9\% & 60\% \\
\midrule
\multirow{4}{*}{EarthFormer vs.\ persistence} & 133 & 0.999 & 0.534 & +7.3\% & +13.1\% & $\uparrow$ \\
 & 160 & 0.993 & 0.387 & -0.4\% & +16.9\% & $\uparrow$ \\
 & 181 & 1.004 & 0.337 & +2.4\% & +25.0\% & $\uparrow$ \\
 & 219 & 1.020 & 0.281 & +8.2\% & +31.3\% & $\uparrow$ \\
 
\bottomrule
\end{tabular}
\end{table}

Table~\ref{tab:boot} gives event-level paired bootstrap intervals for the four
arms of the decomposition, \BootN\ replicates over the \BootEvents\ evaluation
events. The resampling is paired. One event-index vector is shared by all four
arms within a replicate, so the interval on the contrast reflects the
between-arm correlation rather than treating the arms as independent samples.
Only the evaluation half is resampled. The calibration half fixes the transform
once, so knob selection cannot leak into the interval.

These intervals condition on the fitted calibration maps and exclude
calibration-fit uncertainty. The earlier-window refit provides a separate
robustness check, rather than an estimate of that uncertainty.

The first contrast of interest is the knob gain of the backbone alone. At the
reference cell it is \BootKnobDiff\ CSI $[\BootKnobLo, \BootKnobHi]$. That is roughly three
quarters of the whole cascade-over-backbone difference, and it grows
monotonically with the threshold (column six) while the residual does not
(column seven), which is where the rising relative-gain reduction in
Table~\ref{tab:contrasts} comes from. In absolute terms, at $\tau = 219$ a free
monotone transform is worth \BootKnobDiff\ CSI to the backbone, and the
architecture it is being credited against is worth \BootHeadDiff.

The second contrast, in the last column of Table~\ref{tab:boot}, is the part of
the cascade's advantage that remains once both arms carry the same transform. At the reference
cell it is \BootHeadDiff\ $[\BootHeadLo, \BootHeadHi]$, $p \BootHeadPRel$. It is
positive and its interval excludes zero at every threshold under the global map. The interval is conditional on that transform and not on the knob family. Under the per-cell member of $\mathcal{G}$ the same residual reads \EightResidK, and under the exact control it reads \ExactRefCalRel\ (Section~\ref{sec:cascast}). The quoted interval applies only to the global-map result,
not to extrema over $\mathcal{G}$. The intervals are narrow
because the resampling is paired. The two arms see the same events, so the
between-arm correlation is differenced out rather than added.

\begin{table}[h]
\caption{Paired bootstrap over the \BootEvents\ evaluation events, \BootN\
replicates, max $16{\times}16$ pooling. Brackets are 95\% intervals on the
paired contrast; the four CSI columns are point estimates. Column six is
what the free knob is worth to the backbone; column seven is what remains of the
cascade's advantage once both arms carry that same transform. Both contrasts
have $p \BootPMax$ at every threshold. Point estimates are those of
Section~\ref{sec:cascast}. The intervals resample per-event counts cached from
an earlier inference run of the deterministic backbone, whose CSI differs from
the current run by at most $1.1\times10^{-4}$ in any cell; the counts of every
other arm are identical in the two runs.}
\label{tab:boot}
\centering
\small\setlength{\tabcolsep}{4pt}
\begin{tabular}{rrrrrll}
\toprule
\multirow{2}{*}{$\tau$} & \multicolumn{4}{c}{CSI (point estimate)} &
  knob gain & residual \\
\cmidrule(lr){2-5}\cmidrule(lr){6-6}\cmidrule(lr){7-7}
 & backbone & {+}cal & cascade & {+}cal &
   backbone{+}cal $-$ backbone & cascade{+}cal $-$ backbone{+}cal \\
\midrule
16 & 0.7850 & 0.7471 & 0.7833 & 0.7833 & -0.0379 [-0.041, -0.035] & +0.0363 [+0.031, +0.041] \\
74 & 0.6654 & 0.6829 & 0.7146 & 0.7151 & +0.0175 [+0.016, +0.018] & +0.0323 [+0.028, +0.036] \\
133 & 0.4442 & 0.4968 & 0.5638 & 0.5574 & +0.0525 [+0.049, +0.055] & +0.0606 [+0.055, +0.066] \\
160 & 0.3013 & 0.3962 & 0.4592 & 0.4550 & +0.0949 [+0.090, +0.100] & +0.0587 [+0.053, +0.065] \\
181 & 0.2536 & 0.3612 & 0.4176 & 0.4134 & +0.1076 [+0.103, +0.112] & +0.0522 [+0.046, +0.058] \\
219 & 0.1467 & 0.2639 & 0.3069 & 0.2978 & +0.1171 [+0.109, +0.125] & +0.0339 [+0.023, +0.044] \\
 
\bottomrule
\end{tabular}
\end{table}

\section{The exact control and the independently released comparisons}
\label{app:exact}

\paragraph{The global map, split by pooling family.} Under max or no
pooling, where the pre-pooling map is exactly a per-arm threshold, it gives
$r = \PaperFamMaxR$ over \PaperFamMaxN\ cells. Under average pooling, where it
is not, it gives $r = \PaperFamAvgR$ over \PaperFamAvgN\ cells, against
$r = \ExactAvgR$ under the exact control. The difference between the two
controls in Section~\ref{sec:contrasts} therefore comes from the average-pooled
cells, which is where the pre-pooling map can reorder blocks.

\paragraph{CasCast under the exact control.} Table~\ref{tab:exactcascast}
gives the cascade-over-backbone contrast at each threshold under max
$16{\times}16$. The calibrated contrast is positive at $\tau=160$ and $181$ and
negative elsewhere. Every interval excludes zero. The sign of the residual
therefore changes with the threshold as well as with the control. The last column is the share of the absolute CSI difference removed, which is \ExactRefRemoved\ at $\tau=219$. As a share of the relative gain, the basis used in Section~\ref{sec:cascast}, the same cell gives \ExactRefShare. The two differ because calibration also raises the CSI of the backbone, which is the denominator of the relative gain.

\begin{table}[h]
\caption{CasCast cascade against its backbone, max $16{\times}16$, under the
exact control. Brackets are 95\% event-bootstrap intervals; the last column is
the share of the raw contrast removed.}
\label{tab:exactcascast}
\centering
\small
\begin{tabular}{rllr}
\toprule
$\tau$ & raw & calibrated & removed \\
\midrule
74 & +0.0479 [+0.0431, +0.0522] & -0.0124 [-0.0151, -0.0097] & 126\% \\
133 & +0.1180 [+0.1107, +0.1251] & -0.0094 [-0.0127, -0.0062] & 108\% \\
160 & +0.1583 [+0.1502, +0.1661] & +0.0108 [+0.0064, +0.0154] & 93\% \\
181 & +0.1640 [+0.1557, +0.1719] & +0.0193 [+0.0140, +0.0246] & 88\% \\
219 & +0.1615 [+0.1480, +0.1739] & -0.0208 [-0.0282, -0.0131] & 113\% \\
 
\bottomrule
\end{tabular}
\end{table}

\paragraph{DiffCast reproduction.} Our reproduction of DiffCast scores between
\DCReproGapLo\ and \DCReproGapHi\ CSI below the published values, using the
authors' own evaluator. Table~\ref{tab:diffcast} gives the threshold-by-threshold
decomposition behind Section~\ref{sec:independent}.

\begin{table}[h]
\caption{DiffCast against its PhyDNet backbone on SEVIR-128, max
$16{\times}16$, under the exact control. Bias columns are the uncalibrated
pooled frequency bias of each arm; brackets are 95\% intervals on the
calibrated contrast; the last column is the share of the raw contrast removed.}
\label{tab:diffcast}
\centering
\small
\begin{tabular}{rrrrlr}
\toprule
$\tau$ & bias backbone & bias DiffCast & raw & calibrated & removed \\
\midrule
133 & 0.18 & 0.81 & +0.3508 & +0.0099 [+0.0042, +0.0157] & 97\% \\
160 & 0.07 & 0.72 & +0.3252 & +0.0421 [+0.0346, +0.0501] & 87\% \\
181 & 0.04 & 0.69 & +0.2575 & +0.0689 [+0.0597, +0.0782] & 73\% \\
219 & 0.01 & 0.78 & +0.1708 & +0.0406 [+0.0238, +0.0554] & 76\% \\
 
\bottomrule
\end{tabular}
\end{table}

\paragraph{WeatherBench~2 and ECMWF settings.} Both settings use public forecasts of 24-hour precipitation initialised at 00~UTC, and we run no model. The observation is ERA5 on the $0.25^\circ$ grid. An event is at least $\tau$\,mm in a grid cell, or in a $4{\times}4$ block under max or average pooling. Counts are weighted by the cosine of latitude. The control is the exact control of Section~\ref{sec:setup}. For each system, lead time, pooling and threshold, it sets the forecast threshold so that the forecast event frequency on the calibration days equals the observed one. Even-indexed initialisation dates are the calibration days and odd-indexed dates are the evaluation days. Intervals come from a bootstrap over evaluation days in blocks of four. On WeatherBench~2 the systems are GraphCast initialised from ERA5, GraphCast initialised from IFS~HRES, FuXi, IFS~HRES and the IFS ensemble mean. The year is 2020, with 175 calibration and 175 evaluation days. The lead times are 24, 72 and 120\,h and the thresholds are 1, 10, 25 and 50\,mm, so ten pairs at 36 cells give the \WBN\ contrasts. The rank correlation is $\rho = \WBRho$. On ECMWF operational output the systems are AIFS Single, IFS~HRES, and the first perturbed member of each of the AIFS and IFS ensembles. The period is 28 August 2025 to 11 May 2026, with 129 calibration and 128 evaluation days. The lead times are 24, 72 and 144\,h and the thresholds are 1, 5, 10, 20 and 50\,mm, so six pairs at 45 cells give the \ECN\ contrasts. The rank correlation is weaker here, $\rho = \ECRho$. Most frequency biases in this setting are close to one, so most calibration gaps are small. The cells quoted in Section~\ref{sec:independent} use no pooling. The Peirce skill score is the hit rate minus the probability of false detection, and it is the score ECMWF reports for AIFS against IFS \citep{moldovan2025update}. On that score the contrast quoted in Section~\ref{sec:independent} goes from \ECPssSeventyTwoRaw\ to \ECPssSeventyTwoCal.

\paragraph{Predicting the movement out of sample.} Fitting the distortion on the
calibration gap over the \ContrastNArms\ SEVIR arms alone gives
a line with intercept \TTIntercept\ and slope \TTSlope, with residual standard deviation \TTResidSD.
Applied to DiffCast and PreDiff, which are not in the fit, the line places
\TTPDInside\ of the \TTPDN\ PreDiff cells inside the 95\% prediction band,
with a mean absolute error of \TTPDMAE\ CSI, and reproduces the DiffCast
movement with a rank correlation of \TTDCRho\ over \TTDCN\ contrasts among DiffCast, its PhyDNet backbone and persistence. The sign is correct in \TTDCSign\ of the \TTDCNBig\ contrasts whose calibration gap exceeds $0.1$ in absolute value. Inside the range of the fit, \TTDCInside\ of the \TTDCNInRange\ DiffCast contrasts fall within the band. Where PhyDNet
almost never forecasts high values the gap runs past the range of the fit, as
far as \TTDCMaxG, and the straight line overshoots because CSI is bounded, but
the rank and the sign still hold. The movement under the control can therefore
be predicted from the frequency biases alone.

\section{Fitting the map per lead time}
\label{app:lead}

A stricter version of the control fits the map separately at each lead time.
Contrary to our expectation, this raises the residual instead of lowering it.

\paragraph{Status of the per-lead family.} $\mathcal{G}$ is a set of maps of a
scalar, so a horizon-indexed family $\{g_t\}$ is not a member of it. Applied to
the whole field, it can send a pixel at lead $3$ above one at lead $9$ that it had
been below. Its use here rests on a narrower property. Pooling is
spatial, thresholding is pointwise, and the contingency tables are summed across
frames, so every quantity we score is a function of each frame separately and the
cross-frame reordering is invisible to all of them. Within a frame the family
reorders nothing, and each $g_t$ is fitted on the observed marginal at lead $t$
alone, so it is unpaired in the same sense as the rest of the paper. At these
cells it is therefore as free as a member of $\mathcal{G}$, although it is not
one.

\paragraph{Two budgets.} Each $g_t$ is estimated from a twelfth as many pixels as
the global map, and the upper percentiles that decide a $\tau = 219$ cell from a
twelfth as many exceedances, so a loss could be estimation noise rather than the
family. We separate the two. Under the \emph{matched} budget every map is fitted
from the same calibration events, which is the like-for-like comparison. The data are the same and the family is larger. Under the \emph{equal} budget each $g_t$ is given as many
pixels as the global map by drawing more calibration events. The global map is
fitted from a fixed event sample in both, so its column is identical across the
two and the per-lead columns are comparable.

\paragraph{Result.} On the deterministic backbone at the reference cell the
knob gain falls from \LeadGainG\ under the global map to \LeadGainLm\ per lead at
the matched budget and \LeadGainLe\ at the equal budget. Decomposing the cascade
contrast, whose raw gain is \LeadRaw, the relative-gain reduction falls from
\LeadErasedG\ to \LeadErasedLm\ and then to \LeadErasedLe. The smaller reduction persists under the equal-pixel-budget fit. This comparison
does not separately identify family effects and estimation error.
Both cascade arms here are the authors' released checkpoints run through their
released sampler at the same guidance weight and step count. What differs is the
latent-noise realisation, which this pipeline cannot fix exactly because the
conditioning latent is a draw from the VAE posterior through an unseeded
generator. One reproduces the published sampling scheme, one draws noise
independently per event. Their raw gains, \LeadRaw\ and \LeadRawb, straddle the
\EightReported\ of the arm scored in Section~\ref{sec:cascast}, and their relative-gain
reductions under the global map, \LeadErasedG\ and \LeadErasedGb, straddle its
\EightShareQ. Per lead they read \LeadErasedLm\ and \LeadErasedLmb\ at the
matched budget and \LeadErasedLe\ and \LeadErasedLeb\ at the equal budget, so
neither the direction nor its size turns on which sample we score.

\begin{table}[t]
\caption{Pooled frequency bias and $\csi$ by lead time at max $16{\times}16$,
$\tau = 219$, for CasCast's deterministic backbone. ``id'' is uncalibrated,
``G'' the single global quantile map, ``L'' the map fitted per lead time at the
equal-pixel budget. The global map overshoots at the first lead and
under-corrects at the last; the per-lead map is flatter in bias and scores higher at long
leads, but the tables are summed across leads and the early leads carry the
hits.}
\label{tab:lead}
\centering
\small
\begin{tabular}{rrrrrrr}
\toprule
& \multicolumn{3}{c}{pooled bias} & \multicolumn{3}{c}{$\csi$} \\
\cmidrule(lr){2-4}\cmidrule(lr){5-7}
lead & id & G & L & id & G & L \\
\midrule
1 & 0.572 & 1.158 & 0.703 & 0.5000 & 0.5952 & 0.5605 \\
2 & 0.397 & 0.882 & 0.539 & 0.3466 & 0.5208 & 0.4344 \\
3 & 0.262 & 0.652 & 0.435 & 0.2282 & 0.4167 & 0.3324 \\
4 & 0.187 & 0.508 & 0.406 & 0.1604 & 0.3279 & 0.2894 \\
5 & 0.144 & 0.382 & 0.367 & 0.1209 & 0.2559 & 0.2485 \\
6 & 0.117 & 0.314 & 0.356 & 0.0947 & 0.2084 & 0.2257 \\
7 & 0.096 & 0.265 & 0.352 & 0.0752 & 0.1668 & 0.2043 \\
8 & 0.075 & 0.212 & 0.301 & 0.0572 & 0.1385 & 0.1722 \\
9 & 0.064 & 0.184 & 0.305 & 0.0477 & 0.1159 & 0.1672 \\
10 & 0.057 & 0.155 & 0.305 & 0.0426 & 0.0963 & 0.1547 \\
11 & 0.045 & 0.128 & 0.303 & 0.0349 & 0.0791 & 0.1496 \\
12 & 0.032 & 0.103 & 0.316 & 0.0228 & 0.0677 & 0.1396 \\
 
\bottomrule
\end{tabular}
\end{table}

\paragraph{Source of the global map's higher score.} Table~\ref{tab:lead}
shows the tradeoff across horizons. At the first lead, the global map produces
pooled bias above one and higher CSI than the per-lead map. At the last lead,
the per-lead map roughly doubles CSI relative to the global map. Summing
contingency tables across horizons weights these changes by their event counts.
The resulting pooled CSI favours the global map. Thus horizon aggregation,
like spatial pooling, affects the reported comparison.

\paragraph{Implications for the main result.} The relative-gain reduction under the global map is unaffected. Because per-horizon marginal matching
raises the residual, the global result is not conservative along this axis. Appendix~\ref{app:knobsevir} finds the opposite
direction for a bounded shift/gain grid under both selection rules. These are
sensitivity results for the tested procedures. The range in
Section~\ref{sec:cascast} is the range across its three reported controls,
not a bound over $\mathcal{G}$ or its horizon-specific extensions.

\section{Joint sensitivity to transform family and selection rule}
\label{app:knobsevir}

Appendix~\ref{app:knobrule} examines how the selection rule affects the knob
gain, mainly on crowd density, and also discusses SEVIR.
Appendix~\ref{app:lead} found a per-horizon transform producing a smaller
relative-gain reduction than its pooled-horizon counterpart. Both point to the same gap. The knob gain
is a function of the family the knob is drawn from and of the rule that picks a
member, and the paper reports one of each. This crosses them at the cell
the main results are quoted at.

\paragraph{Terminology.} ``Global'' is reserved throughout the paper for the
single quantile map held fixed across every cell, which brings the backbone to
\EightDetQ. That map does not appear in this appendix. The rows here are the discrete
family of shifts and gains of Section~\ref{sec:setup}, selected per cell.
\emph{Pooled-horizon} selects one member per cell from a contingency table summed
over lead times, which is the per-cell knob of the main text. \emph{Per-horizon}
selects one member per (cell, lead).

\paragraph{Agreement with the main audit.} Under
pooled-horizon selection by bias, which is the per-cell knob of
Section~\ref{sec:cascast}, the backbone goes from \KnobDetUncal\ uncalibrated to
\KnobDetPooledBias\ at the reference cell. This matches the main audit's per-cell
value, \EightDetK, to four figures, although this harness accumulates a
separate, lead-resolved contingency table.

\begin{table}[h]
\caption{The knob family crossed against the selection rule, at max
$16{\times}16$, $\tau = 219$, decomposing the cascade contrast whose raw gain is
\KnobRaw. Both rules see the calibration half only. ``per-horizon'' selects a
member per (cell, lead) and is not a member of $\mathcal{G}$; it is
reported as an additional sensitivity result. The range in
Section~\ref{sec:cascast} compares its three specified controls only.}
\label{tab:knobsevir}
\centering
\small
\begin{tabular}{llrrrr}
\toprule
family & rule & backbone & cascade & residual & rel.-gain reduction \\
\midrule
pooled-horizon & bias & 0.3201 & 0.3059 & -4.4\% & 104.0\% \\
pooled-horizon & val-CSI & 0.3201 & 0.3123 & -2.4\% & 102.2\% \\
per-horizon & bias & 0.3366 & 0.2990 & -11.2\% & 110.0\% \\
per-horizon & val-CSI & 0.3413 & 0.3086 & -9.6\% & 108.6\% \\
 
\bottomrule
\end{tabular}
\end{table}

\paragraph{Effect of the family and of the rule.} Read down
Table~\ref{tab:knobsevir}, the two rules differ by less than the two families do.
At the reference cell both rules select the same member, \KnobDetKnob, so the
pooled-horizon rows differ only through the cascade arm, and selecting on
$\csi$ is not uniformly better. Appendix~\ref{app:knobrule} gives the cell where
it loses out of sample. Crossing to per-horizon moves the backbone from
\KnobDetPooledBias\ to \KnobDetPerhCsi\ and the residual from \KnobResidPooled\
to \KnobResidPerhCsi. The direction differs from
Appendix~\ref{app:lead} under both selection rules. The per-horizon quantile map
and the bounded shift/gain grid impose different calibration constraints and
select different operating points. This does not establish an inherent inability
of shifts or gains to reproduce the quantile map's exceedances. At a single
threshold under max pooling, an unrestricted shift can reproduce the same upper
set, subject to boundary and tie conventions. The distinction here concerns the
finite candidate grid and the fitting procedures tested.

\paragraph{Interpretation.} A per-horizon member can
send a pixel at one lead above a pixel at another that it had been below, so like
the per-horizon map of Appendix~\ref{app:lead} it is not a member of
$\mathcal{G}$ as Section~\ref{sec:setup} defines it. It is as free as one at
these cells, for the same reason. Pooling is spatial, thresholding is pointwise,
and the tables are summed across frames, so no scored quantity sees the
cross-frame reordering. The second cascade sample of Appendix~\ref{app:lead} agrees and is marginally
stronger. Against its raw \KnobRawb\ the pooled-horizon residual reads
\KnobResidPooledb\ and the per-horizon one \KnobResidPerhCsib, corresponding to
relative-gain reductions of \KnobErasedPooledb\ and \KnobErasedPerhCsib. The two samples straddle the \EightResidK\ pooled-horizon residual reported in
Section~\ref{sec:cascast}, as they straddle \EightShareQ\ under the global map,
so the crossing is not a property of which sample of the cascade is scored.

Section~\ref{sec:cascast} reports the global, per-cell and exact controls. The per-horizon results are
additional sensitivity checks. Both per-horizon rows leave a negative residual.
The bias-selected row uses unpaired marginal information. The validation-CSI
row uses paired labels and is a separate robustness check. Neither provides a
certified bound over the transform family. The range in the abstract covers the three controls of Section~\ref{sec:cascast} only.

\section{Alternative verification scores}
\label{app:relwork}

Section~\ref{sec:related} claims that changing the score does not remove the
confound studied here. Because that claim motivates our choice to transform the
forecast, we examine the alternatives in turn. Throughout, $B$ denotes frequency bias and $s$ the base rate of
the thresholded event. Our reference cell sits at small $s$.

\paragraph{A published post-processor moving the same cell.} PostCast
\citep{gong2025postcast} is the clearest external evidence that this operating
point is movable after the forecaster is frozen. It is an unsupervised
deblurring stage applied to a finished nowcast, and on SEVIR at the extreme
threshold it takes TAU from $\csi = 0.008$ to $0.043$ unpooled, $0.014$ to
$0.074$ under $4{\times}4$ max pooling, and $0.028$ to $0.163$ under
$16{\times}16$. The absolute movement grows with the pooling scale, $+0.035$
unpooled against $+0.135$ at $16{\times}16$, consistent with the bias--gain relationship
of Section~\ref{sec:mechanism} in independently published results. The relative gain
is roughly flat near $5\times$, so the absolute room to move is largest in the
pooled cell. No frequency bias is reported for any arm.

The PostCast transform, however, is a diffusion model conditioned
on the blurry input, so it reorders pixels and sits outside $\mathcal{G}$. We
cannot say its gain is free, and some of it may reflect placement. Its operating
cell nevertheless matches ours. Under the standard SEVIR VIL encoding
$R(x) = \exp((x - 83.9)/38.9)$ for $x > 18$, their $\tau = 32.24$\,kg\,m$^{-2}$
is byte $219$ to three figures, and they pool $16{\times}16$ with a max. What
differs is the grid (their SEVIR is resized to $256{\times}256$, so a block
covers more ground than ours), and that, with checkpoints we could not
confirm as released, is why their artefacts could not enter the pool of
Section~\ref{sec:contrasts} without retraining. These results nonetheless indicate the scale of the effect. A post-hoc operation on a frozen forecast
multiplies a reported extreme-threshold pooled CSI by about six, and the
comparison is published without a frequency-bias diagnostic or a symmetric
marginal control to assess that contribution.

\paragraph{Bias-adjusted threat score.} \citet{mesinger2004bias,mesinger2008bias}
estimate the threat score at unit bias using an assumed relationship between
hits and forecast frequency. In the 2008 formulation, the increase in hits per
additional false alarm is proportional to the remaining unhit area. The
adjustment estimates a counterfactual hit count from the original contingency
table. Our audit instead applies a held-out transform to the forecast values
and measures the resulting table. Two forecasts with identical original
contingency counts can produce different counts after calibration, because
those counts do not retain the ordering of values within the thresholded
categories. A bias-adjusted score therefore cannot, in general, recover the
contrast under our specified transform. It answers a complementary question.
We do not claim that its assumptions fail on the fields evaluated here.

\paragraph{Chance-corrected scores.} ETS subtracts the hits a random forecast of
the same frequency would obtain. That corrects for chance, not for bias, and the
two are distinct. A forecast can be perfectly chance-corrected and still be
scored at a bias far from one. \citet{hogan2010equitability} sharpen the point,
showing that ETS satisfies neither of Gandin and Murphy's requirements for
equitability, and satisfies the first only in the limit of infinite sample size.
Appendix~\ref{app:ets} reports the decomposition of Section~\ref{sec:cascast}
re-run under ETS. HSS is a strictly increasing function of ETS, so every
contrast sign and every reversal is identical under the two by identity rather
than by experiment, and we do not report it separately.

\paragraph{Rare-event scores.} The extreme dependency score
\citep{stephenson2008eds} and the odds-ratio skill score
\citep{stephenson2000odds} were built for the regime we operate in, where
linear measures degenerate toward a constant as $s \to 0$. The extremal
dependence indices \citep{ferro2011edi} repair EDS's own hedgeability and
base-rate dependence. These scores improve on CSI for rare events, and reporting
SEDI alongside CSI would benefit nowcasting evaluation. However, they do not
remove the sensitivity examined here.
\citet{hogan2010equitability} show that ORSS and the symmetric extreme
dependency score are more strongly inequitable than ETS for rare events
(at a base rate of 2\% with 1000 samples, an unbiased random forecast scores
about $-0.5$), so at small $s$ the substitution moves the bias sensitivity
around rather than removing it. Two further points are specific to our setting.
First, none of these scores is what the benchmark reports. The claims we audit
are stated in pooled CSI, and an audit conducted with a different score would
address a different question. Second, and more basic, the mechanism we document
is upstream of the choice of contingency-table score, because it operates on
which cells enter the table at all. Pooling is applied to the field before any
score sees it.

\paragraph{Neighbourhood and object-based alternatives.} FSS
\citep{roberts2008fss} is the neighbourhood literature's own answer, and it
carries the same vulnerability. Its large-neighbourhood asymptote is determined
by frequency bias rather than by placement \citep{necker2024pfss}. Feature- and
object-based methods provide separate diagnostics of structure, amplitude, and
location. MODE uses object matching \citep{davis2006mode}, whereas SAL summarises
identified objects without one-to-one matching \citep{wernli2008sal}. These
methods require choices about object identification and, for MODE, matching.
Their sensitivity to amplitude is not evaluated here. They complement an audit
of results already stated in pooled CSI, but answer a different question.

\paragraph{A control on the forecast.} Every alternative above modifies the
score. Our control modifies the forecast, using a
map that is monotone, fitted without reference to the test targets under the
protocol of Section~\ref{sec:setup}, and available to every arm at no training
cost. Its interpretation is correspondingly narrow. It does not say what the
better model is, only how much of a reported gap survives when the free part of
the gap is given to everyone. This narrower claim is the one that the released
checkpoints can support.

\end{document}